\documentclass[11pt]{article}
\usepackage[margin=.5in]{geometry}
\usepackage{fullpage}
\usepackage{multirow}
\usepackage{amsgen,amsmath,amstext,amsbsy,amsopn,amssymb, amsthm}
\usepackage{bm}
\usepackage{url}
\usepackage{graphicx}
\usepackage{caption}
\usepackage{subcaption}
\usepackage{algorithm}
\usepackage{tablefootnote}
\usepackage{threeparttable}

\usepackage[mathscr]{eucal}

\usepackage{amsbsy}

\usepackage{comment}
\usepackage{enumitem}
\usepackage{booktabs}
\usepackage{algpseudocode}
\usepackage{longtable}
\usepackage{mleftright}

\usepackage{thmtools}

\usepackage{fixme}
\fxsetup{
  status=draft,
  theme=color,
  silent,
}
\fxsetup{marginface=\linespread{1}\tiny}
\FXRegisterAuthor{tk}{tke}{TK} 
\FXRegisterAuthor{rt}{rte}{RT}
\FXRegisterAuthor{az}{aze}{AZ}

\usepackage{hyperref} % Should be last
\hypersetup{
	colorlinks=true,
	linkcolor=blue,
	filecolor=blue,      
	urlcolor=red,
	citecolor=blue,
}
\usepackage[capitalize,nameinlink]{cleveref} % except for this one

\DeclareMathAlphabet\mathbfcal{OMS}{cmsy}{b}{n}

\newcommand{\argmin}{\operatornamewithlimits{argmin}}

\usepackage{color}

\newcommand{\be}{\begin{equation}}
\newcommand{\ee}{\end{equation}}

\newcommand{\F}{\mathrm{F}}

\newcommand{\bcX}{{\boldsymbol{\mathscr{X}}}}

\newcommand{\bcY}{{\boldsymbol{\mathscr{Y}}}}

\newcommand{\bcZ}{{\mathbfcal{Z}}}

\newcommand{\T}{^\top}

\newcommand{\RR}{{\mathbb{R}}}

\newtheorem{theorem}{Theorem}
\newtheorem{remark}{Remark}
\newtheorem{Example}{Example}

\newtheorem{proposition}{Proposition}

\algrenewcommand\algorithmicensure{\textbf{Output:}}
\algrenewcommand\algorithmicrequire{\textbf{Input:}}

\allowdisplaybreaks

\title{Tensor Decomposition with Unaligned Observations}

\author{Runshi Tang\footnote{Department of Statistics, University of Wisconsin-Madison; Email: \texttt{rtang56@wisc.edu}}, ~ Tamara Kolda\footnote{MathSci.ai; Email: \texttt{tammy.kolda@mathsci.ai}}, ~ and ~ Anru R. Zhang\footnote{Departments of Biostatistics \& Bioinformatics and Computer Science, Duke University; Email: \texttt{anru.zhang@duke.edu}; supported in part by the NSF Grant CAREER-2203741.}}

\date{\today}

\begin{document}
	
\maketitle

\begin{abstract}
This paper presents a canonical polyadic (CP) tensor decomposition that addresses unaligned observations. The mode with unaligned observations is represented using functions in a reproducing kernel Hilbert space (RKHS). We introduce a versatile loss function that effectively accounts for various types of data, including binary, integer-valued, and positive-valued types. Additionally, we propose an optimization algorithm for computing tensor decompositions with unaligned observations, along with a stochastic gradient method to enhance computational efficiency. A sketching algorithm is also introduced to further improve efficiency when using the $\ell_2$ loss function. To demonstrate the efficacy of our methods, we provide illustrative examples using both synthetic data and an early childhood human microbiome dataset.
\end{abstract}

%%%%%%%%%%%
\section{Introduction}\label{sec:intro}
%%%%%%%%%%%

\begin{figure}[!ht]
	\centering
        \hspace{.5cm}
	\begin{subfigure}[t]{0.3\textwidth}
		\includegraphics[width=\textwidth]{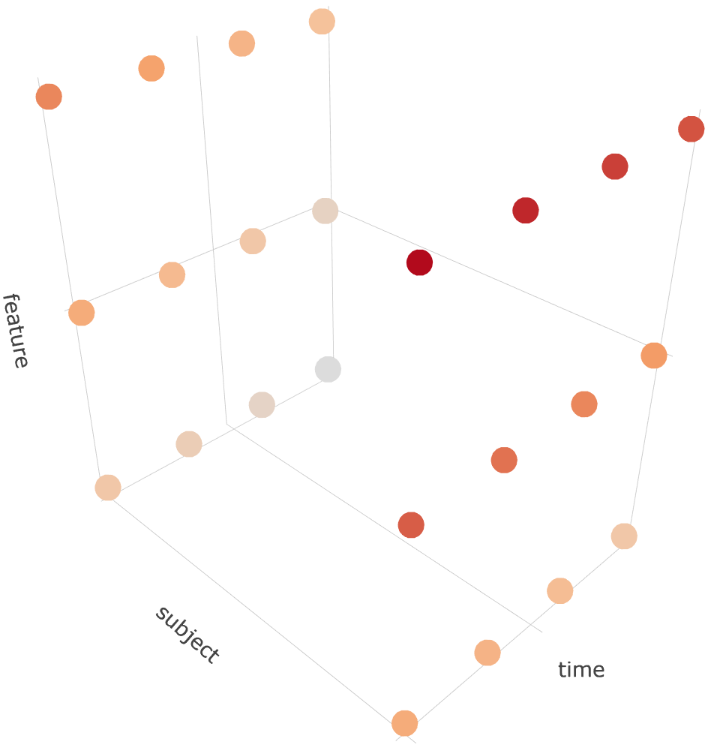}
		\caption{Tabular tensor}\label{fig:tabular-functional-tensor-a}
	\end{subfigure}
        \begin{subfigure}[t]{0.3\textwidth}
		\includegraphics[width=\textwidth]{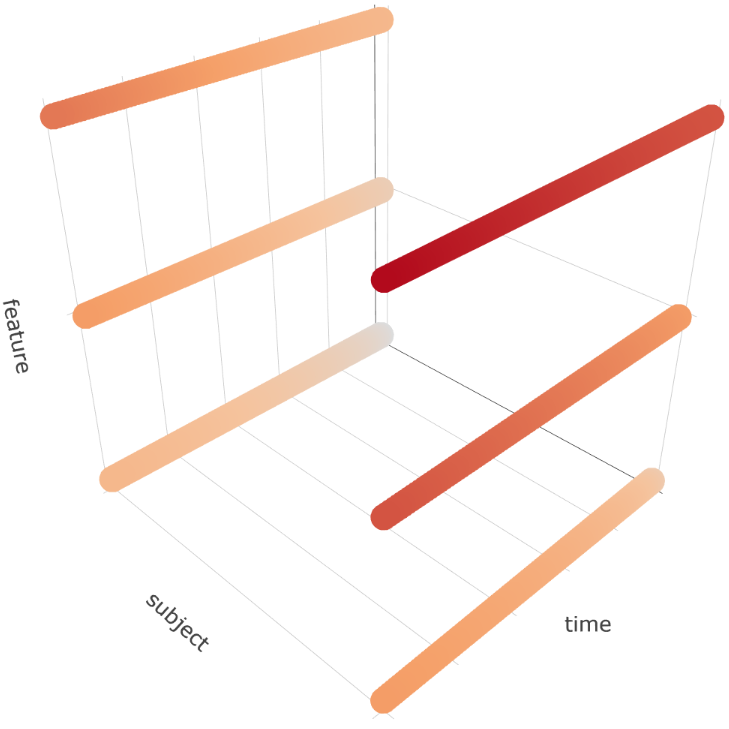}
		\caption{Functional tensor}\label{fig:tabular-functional-tensor-b}
	\end{subfigure}
	\begin{subfigure}[t]{0.3\textwidth}
		\includegraphics[width=\textwidth]{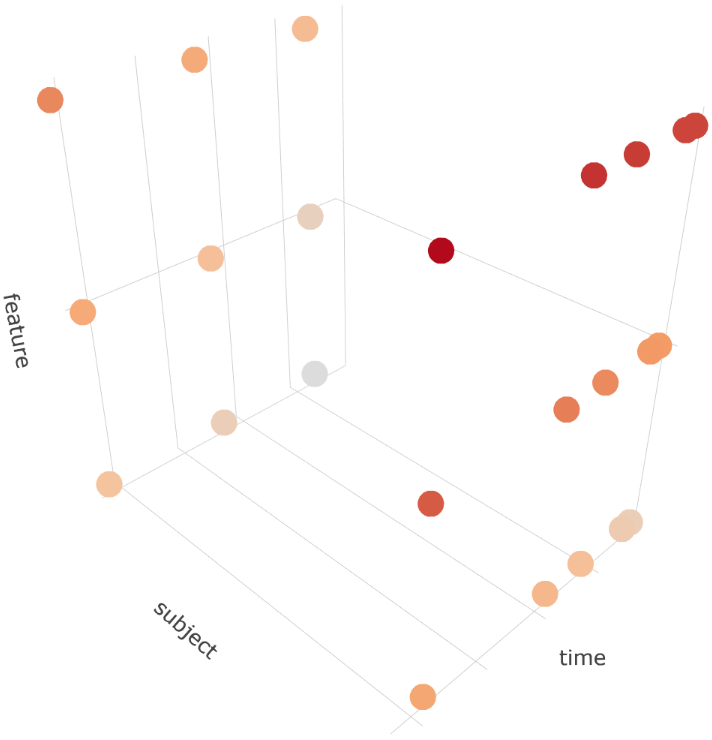}
		\caption{Unaligned observations of functional tensor}\label{fig:tabular-functional-tensor-c}
	\end{subfigure}
        \hspace{.5cm}
     \caption{Comparison of tabular and functional tensors. The color represents the tensor entry value from 0 (white) to 1 (red).}
     \label{fig:tabular-functional-tensor}
\end{figure}

Tensor data analysis is a cutting-edge field that resides at the intersection of mathematics, statistics, and data science. 
It specializes in handling datasets that can be formatted into three or more directions. 
In comparison to traditional vector or matrix data, tensors efficiently represent high-order information, capturing intricate relationships and patterns across multiple modes and dimensions simultaneously \cite{sidiropoulos2017tensor}. 
A pivotal step of tensor analysis is decomposition, which aims to compute a low-rank approximation of a given tensor. 
In a variety of applications, the input tensor is tabular, i.e., a multi-way array (\cref{fig:tabular-functional-tensor-a}). 
In essence, an order-3 tabular tensor $\mathcal{T}$ taking values is equivalent to a mapping: 
\begin{equation*}
    \begin{split}
        [n] \times [p] \times [q] & \to \mathbb{R}\\
        (i,j,k)& \mapsto \mathcal{T}_{ijk}.
    \end{split}
\end{equation*}

However, in another class of applications, tensor data can be presented as a collection of functions. 
For instance, consider the situation where we observe $\bcY_{ij}(t)$ for feature $j$ of subject $i$ at time $t$ (see \cref{fig:tabular-functional-tensor-b}).
\begin{figure}[htbp]
        \centering
		\includegraphics[width=\textwidth]{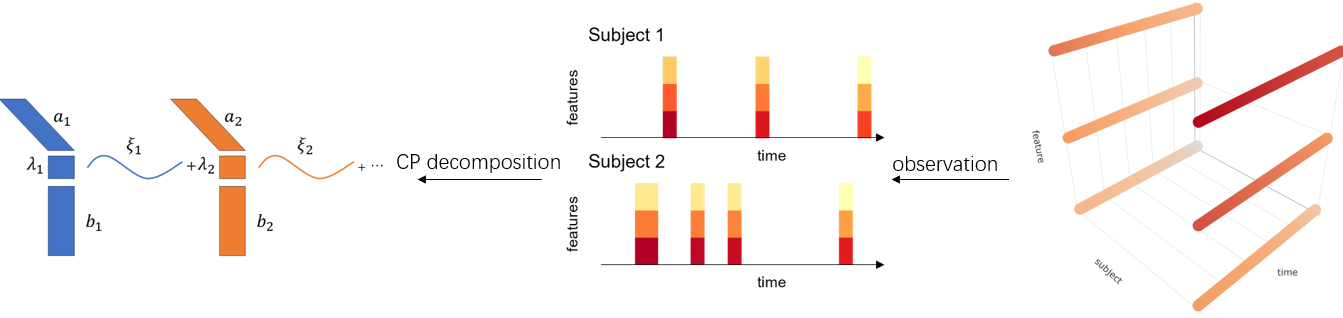}
	\caption{Illustration of tensor decomposition with unaligned observations}
	\label{fig_Illustration}
\end{figure}
In such cases, the tabular tensor model encounters two limitations. First, the order of indices within the tabular tensor is often treated as exchangeable, which ignores the sequential structures of the functional mode. 
Secondly, the model assumes that observations are consistently aligned across all indices. 
This assumption often fails to hold, particularly in the time mode of multivariate longitudinal studies. 
For example, nonhospitalized patients may have physical exams on different days. 
In these situations, subject $i$ may be measured at a set of time points, denoted as $T_i$, and these sets may vary among different subjects (as illustrated in \cref{fig:tabular-functional-tensor-c}), and we describe this as \emph{tensor with unaligned observations}.

Recently, \cite{han2021guaranteed} proposed a statistical model named \emph{functional tensor singular value decomposition} (FTSVD). This model extended the CP decomposition to functional tensors and proposed a power iteration algorithm to obtain an approximate decomposition: $\bcX_{ij}(t) = \sum_{r=1}^R (a_r)_i\cdot (b_r)_j\cdot \xi_r(t) + \bcZ_{ij}(t), i\in[n], j\in[p], t\in T$, where $a_r, b_r$ are vectors, $\xi_r$ are functions, and $\bcZ_{ij}(t)$ refers to random noise. The authors also proved upper bounds on the estimation error for their estimator. However, the framework proposed in \cite{han2021guaranteed} has two limitations. First, the proposed estimation procedure requires observational time points to be aligned among different subjects, i.e., $T$ must be the same across all subjects $i$. Second, the model assumes an additive relationship between the low-rank signal and Gaussian random noise, which may not be appropriate for certain types of data, such as discrete-valued data. We will demonstrate the importance of addressing these limitations through real-world research examples, including the following two:
\begin{Example}\label{example_clinical_trail}
    In the study by \cite{towe2023longitudinal}, patients are scheduled for clinical tests every 30 days over 5 visits. While ideal scheduling allows for tabular tensor organization of the data, real-world attendance varies, causing inconsistencies in time intervals between visits. The measured scores at each visit follow binomial or Poisson distributions. To apply the algorithm from \cite{han2021guaranteed} or any tabular tensor decomposition method, observations must be aligned to fixed 30-day intervals, assuming the $i$th visit occurs exactly on the $30 \times (i-1)$st day for all patients. This alignment introduces bias. Moreover, a transformation is needed to handle the binomial or Poisson-distributed measurements in \cite{han2021guaranteed}, adding further bias.
    \end{Example}

\begin{Example}\label{example_ecam}
    The Early Childhood Antibiotics and the Microbiome (ECAM) study \cite{bokulich2016antibiotics} involves 42 infants with irregularly timed fecal microbiome gene sequence counts from birth to two years. Due to the inconsistent timing, the algorithm from \cite{han2021guaranteed} and other tabular tensor decomposition methods cannot be directly applied. An alternative is to calculate mean or median values at fixed intervals (e.g., monthly) for each infant, but this results in information loss. Additionally, a transformation is needed to handle the counting data in \cite{han2021guaranteed}.

    %The Early Childhood Antibiotics and the Microbiome (ECAM) study \cite{bokulich2016antibiotics} includes data from 42 infants, with multiple fecal microbiome measurements of gene sequence counts taken from birth through the first two years of life. The timing of these measurements is irregular, with no consistent period or pattern. As a result, the algorithm from \cite{han2021guaranteed} or other tabular tensor decomposition methods cannot be directly applied to this dataset. An alternative approach is to calculate the mean or median at fixed intervals (e.g., monthly) for each infant and each microbiome measurement, which will result in a loss of information. Additionally, to properly handle the counting data, a transformation would be required for \cite{han2021guaranteed}.
\end{Example}

%In this context, we treat the mode with unaligned observations as functional. We wish to compute a canonical polyadic (CP)-like tensor decomposition that breaks down high-dimensional data represented as tensors into a set of simpler, lower-dimensional, and more interpretable components. While a substantial body of literature has focused on decomposition techniques for tensors with structured tabular formats \cite{de2000multilinear,hong2020generalized,kolda2009tensor,oseledets2011tensor,zhang2018tensor}, there has been limited exploration regarding tensor decomposition methods that include functional modes within their scope.
To address these limitations, this paper introduces a new tensor decomposition framework called {\it tensor decomposition with unaligned observations}. Suppose we observe $\bcX_{ij}(t)$, where $i\in[n], j\in[p], t \in T_i$ from an overall tensor $\bcX\in \mathbb{R}^{n\times p \times T}$, noting that $T_i$ may differ across different subjects $i$. We propose a new method to compute vectors $a_r\in \mathbb{R}^n, b_r\in \mathbb{R}^p$, and functions $\xi_r$ such that $\bcX$ and $\sum_{r=1}^R a_r \circ b_r\circ \xi_r$ align based on specific criteria. We specifically propose to compute: 
\begin{equation}\label{optimization_problem_generalized_sec1}
    \{\hat{a}_r, \hat{b}_r, \hat{\xi}_r\}_{r =1}^R = \argmin_{\substack{(a_r, b_r, \xi_r) \in \Phi \\ {r = 1,\ldots,R}}} \frac{1}{|\Omega|}\sum_{i=1}^n\sum_{j=1}^p \sum_{t\in T_i} f\left(\sum_{r=1}^R (a_r)_i\cdot (b_r)_j\cdot \xi_r(t), \bcX_{ij}(t)\right).
\end{equation}
Here, $\Phi$ represents some feasible set that promotes structures such as smoothness or nonnegativity, which will be specified later. While a canonical choice of $f$ could be the $\ell_2$ loss, i.e., $f(a, b) = (a-b)^2$, we can also employ a more versatile loss function $f$ similar to the concept of the \emph{generalized} CP (GCP) decomposition \cite{hong2020generalized}, which effectively accounts for different types of data, including binary, integer-valued, and positive-valued types. See \cref{fig_Illustration} for a pictorial illustration of our framework.

Our framework is particularly useful in the context of longitudinal multivariate analysis, where longitudinal data containing multiple features from different subjects are common across a range of applications \cite{verbeke2014analysis}.  These data can be organized into a tensor consisting of two tabular modes that represent subjects and features, alongside a functional mode that represents unaligned time points. Examples of such applications include \cref{example_clinical_trail} and \cref{example_ecam}.

The irregularly observed time points in the unaligned mode can lead to increased computational time requirements for the method, resulting in an optimization problem with much higher dimensionality. To address this issue, this paper introduces a stochastic gradient descent technique to solve \eqref{optimization_problem_generalized_sec1} and a sketching technique when $f$ is specifically the $\ell_2$ loss. By incorporating these techniques, we significantly reduce the computation time while maintaining the desired level of accuracy in simulation studies. 

Finally, we test and compare the performance of our proposed algorithms in simulation studies and a microbiology dataset of early childhood human microbiomes, which shows the applicability of our proposed approaches. We compare our approach with some benchmark methods in the literature, including functional tensor singular value decomposition, standard CP decomposition, and standard GCP decomposition, which demonstrates the advantage of our approach. 

%%%%%%%%%%%
\subsection{Literature Review}\label{sec:literature-review}
%%%%%%%%%%%

For a broad introduction, readers are referred to surveys on tensor decomposition
\cite{kolda2009tensor,sidiropoulos2017tensor}. 
Numerous methodological variations have emerged in the literature. 
For example, a variation of the classical CP decomposition, called CANDELINC was proposed in \cite{douglas1980candelinc}, which constrains the column spaces of one or more of the factor matrices. 
In \cite{cohen2017dictionary}, the authors considered a case that enforces one factor to belong exactly to a known dictionary. 
\cite{sorber2015structured} introduced a more generalized framework with the least squares loss. 
\cite{battaglino2018practical,vervliet2015randomized} proposed randomized CP decomposition methods tailored for the fast computation of large-scale tabular tensors. 
The paper~\cite{hong2020generalized} introduced GCP, which allows for different loss functions in computing CP decomposition and proposed an algorithm based on gradient descent, and \cite{kolda2020stochastic} incorporated stochastic gradient descent into the aforementioned GCP decomposition framework. 

Other work on functional CP tensor decomposition includes \cite{BeMo02,BeMo05,ChLeNoRa15, de2011blind, timmerman2002three}. Specifically, \cite{BeMo02} introduced a CP-decomposition-type representation of high-dimensional functions that expresses the $d$-variate function as the sum of products of $d$ univariate functions and showed that the multiparticle Schr\"odinger operator and inverse Laplacian can be efficiently represented by this form. They treated every mode as functional, rather than just one mode. In \cite{de2011blind}, the authors considered the tensor with functional modes generated by exponential polynomials and required the observed tensor to be tabular. \cite{timmerman2002three} introduced smoothness constraints to CP decomposition by assuming one factor matrix could be further decomposed as the product of a B-spline matrix and a weight matrix. They also assume tabular observations. 
{Additionally, \cite{lacroix2020tensor} discussed the CP decomposition on tensors with a temporal mode for link prediction in knowledge base completion.
\cite{sapienza2018non} applied the classical CP decomposition to tensors with an aligned temporal mode to explore hidden correlated behavioral patterns in multiplayer online game data. 
\cite{zhu2017tensor} considered the tucker decomposition applying to the tensor with one functional mode from Magnetic Resonance Image data. }

The reproducing kernel Hilbert space (RKHS) is a fundamental tool in machine learning. 
It offers a powerful mathematical framework for handling complex data and has found widespread applications in various topics. 
RKHS was first introduced in \cite{aronszajn1950theory} as a generalization of the notion of a Hilbert space for functions. 
This concept flourished later with the increasing popularity of machine learning. For those seeking further insights into RKHS, see reference books \cite{kennedy2013hilbert, scholkopf2002learning,shawe2004kernel}.

Sketching is a fundamental concept in the domain of randomized numerical linear algebra, a field that leverages probabilistic algorithms for various linear algebra computations. It can accelerate the computation speed of tasks like matrix multiplication and least squares problems \cite{avron2010blendenpik,drineas2011faster, mahoney2011randomized,vervliet2015randomized}. 
Comprehensive surveys and books on this subject, such as \cite{mahoney2011randomized,martinsson2020randomized,woodruff2014sketching}, provide in-depth insights and resources.
Recently, sketching has found application in tensor CP decomposition \cite{battaglino2018practical,bharadwaj2023distributedmemory, bharadwaj2023fast, LaKo22, malik2022samplingbased}. 

Stochastic gradient descent (SGD) is a randomized algorithm that substitutes the original full gradient with a random sparse gradient whose expectation equals the original full gradient during each iteration of regular gradient descent.  SGD has grown into a pivotal optimization method in machine learning \cite{murphy2022probabilistic, saad1998online, sra2012optimization}. 
This technique has found applications in both the standard and generalized tensor CP decomposition methods \cite{ge2015escaping,kolda2020stochastic,vervliet2015randomized}.

A more recent study \cite{larsen2024tensor} considers tensor decomposition with unaligned observations in three or more modes, including one continuous mode. In addition, \cite{larsen2024tensor} examines the aligned-observation setting and demonstrates that the computational complexity can be significantly smaller than the unaligned cases. While \cite{larsen2024tensor} focuses on the $\ell_2$ loss and tensors of general order, our work specifically targets order-3 tensors, developing explicit algorithms to handle more general loss functions. Additionally, we elaborate on acceleration schemes, including sketching and stochastic gradient approaches, to efficiently address the high computational complexity introduced by unaligned observations.

%%%%%%%%%%%
\section{Notation and Preliminaries}\label{sec:preliminary}
%%%%%%%%%%%

For any vector \(x = (x_1, \ldots, x_m) \in \mathbb{R}^{m}\), let \(\|x\| = \sqrt{x_1^2 + \ldots + x_m^2}\) be its \(l_2\) norm. 
For any finite set \(S = \{s_1, s_2, \ldots, s_n\}\), \(|S| = n\) denotes its cardinality. 
If \(\xi\) is a function, we denote \(\xi(S) = [\xi(s_1), \xi(s_2), \ldots, \xi(s_n)]^\top \in \mathbb{R}^n\). 
Let \(\xi_1, \ldots, \xi_R\) be \(R\) functions; we denote \(\Xi(S) = [\xi_1(S), \ldots, \xi_R(S)] \in \mathbb{R}^{|S| \times R}\). We use bold uppercase calligraphy letters (e.g., $\bcX, \bcY$) to denote tensors. 
An order-$3$ tabular tensor $\bcX\in \RR^{p_1\times p_2 \times p_3}$ can be viewed as a trivariate function, where $(i_1, i_2, i_2)$ maps to $\bcX_{i_1 i_2 i_3} \in \RR$. 
An order-3 tensor with two tabular modes and one functional mode $\bcX \in \RR^{p_1\times p_2 \times T}$ can be viewed as a map from $(i, j, t)$ to $\bcX_{ij}(t) \in \RR$, where $i\in [p_1], j\in [p_2], t \in T$, and $T$ is some interval in $\RR$. 
The Hadamard product between two matrices $A$ and $B$ with the same size is denoted as $A* B$. For convenience, we also use $*$ to denote the broadcasted element-wise product between a matrix $A = [a_1, \ldots, a_p]$ and a vector $b$: $A*b = [a_1*b, \ldots, a_p*b]$. For a
The CP decomposition of an order-3 tabular tensor is defined as $\bcX = \sum_{i = 1}^r a_i \circ b_i \circ c_i$, where $r$ is the CP rank of $\bcX$ and $a_i \circ b_i \circ c_i$ is an outer product of three vectors $a_i, b_i$, and $c_i$. 
The readers are referred to \cite{kolda2009tensor} for more preliminaries on algebra, operation, and decomposition of tensors. 

Next, we discuss the preliminaries of the reproducing kernel Hilbert space (RKHS). 
We use $\mathcal{L}^2([0,1])$ to denote the space of all square-integrable functions, i.e., 
\begin{displaymath}
\mathcal{L}^2([0,1])=\left\{f:[0,1] \rightarrow \mathbb{R},\|f\|_{\mathcal{L}^2}^2<\infty\right\}, \quad \text {where} \quad\|f\|_{\mathcal{L}^2}:=\left(\int_0^1 f^2(t) d t\right)^{1 / 2}.
\end{displaymath}
For a Hilbert space $\mathcal{H} \subseteq\mathcal{L}^2([0,1])$ associated with the inner product $\langle\cdot, \cdot\rangle_{\mathcal{H}}$ and norm $\|\cdot\|_{\mathcal{H}}$, suppose there is a continuous, symmetric, and positive-semidefinite kernel function $\mathbb{K}:[0,1] \times[0,1] \rightarrow \mathbb{R}_{+}$ that satisfies the following RKHS conditions:
(1) for any $s \in[0,1],
\mathbb{K}(\cdot, s) \in \mathcal{H};$ (2) for each $g \in \mathcal{H}$, $g(t)=\langle g,
\mathbb{K}(\cdot, t)\rangle_{\mathcal{H}}$ for all $t \in[0,1]$. We call the map $\mathbb{K}(\cdot, t): [0,1] \rightarrow \mathcal{H}$ the feature map, and with this feature map, the Representer theorem \cite{kimeldorf1970correspondence, scholkopf2001generalized} states that a large family of optimization problems in $\mathcal H$ with proper regularization admit solutions of the form of a linear combination of feature maps: $f = \sum_{i=1}^{n} \theta_i k(., x_i)$ with samples $x_i$.
{Notably, an RKHS always includes continuous functions, ensuring that point evaluations are well-defined. Furthermore, the pointwise evaluation functionals on the RKHS are continuous.}

This work uses the radial kernel $\mathbb{K}_r(x,y) = \exp(-|x-y|^2)$ and the Bernoulli polynomial kernel
\begin{equation}\label{eq_Bernoulli_ker}
    \mathbb{K}_b(x, y)=1+k_1(x) k_1(y)+k_2(x) k_2(y)-k_4(|x-y|),
\end{equation}
where $k_1(x)=x-.5, k_2(x)=\left(k_1^2(x)-1 / 12\right)/2$, and $k_4(x)=\left(k_1^4(x)-k_1^2(x) / 2+7 / 240\right) / 24$ for any $x \in[0,1]$. $\mathbb{K}_r$ is also known as Gaussian kernel and widely studied in the literature, e.g., \cite{cristianini2000introduction, james2013introduction, shawe2004kernel}. 
As outlined in \cite{gu2013smoothing}, $\mathbb{K}_b$ is the reproducing kernel for the Hilbert space $\mathbb{K}^{2,2}=\left\{f: f^{(r)}\right.$, the $r$-th derivative of $f$, is absolutely continuous, $\left.r=0,1,2 ; f^{(2)} \in \mathcal{L}^2([0,1])\right\}$.
The readers are referred to \cite{aronszajn1950theory, kennedy2013hilbert, scholkopf2002learning,shawe2004kernel} for more discussions on RKHS and their use in function approximation.

Furthermore, we introduce the sketching technique in randomized numerical linear algebra.
In this work, we refer to a random matrix \( M_s \in \mathbb{R}^{k \times n} \) as a {\it sketching matrix} or a {\it row-sampling matrix} if each row of \( M_s \) has exactly one nonzero value of $\sqrt{n/k}$ in a position chosen uniformly at random and zero in all other positions. 
For another matrix $A$ with $n$ rows, the product $M_s A$ can be interpreted as a matrix consisting of $k$ uniformly sampled and resampled rows from $A$. 
This technique is useful in fast approximation of large linear computations \cite{mahoney2011randomized}. 
For instance, given matrices $A\in\RR^{m\times n}$ and $B\in\RR^{p\times n}$, the time cost of computing the matrix product $A\T B$ is on the order of $O(mnp)$ and can be substantial when $n$ is large. 
However, we can approximate this product as $A\T B \approx A\T S\T SB = (M_s A)\T (M_s B)$ without the need to calculate the products $M_s B$ or $M_s A$, as these products can be directly written down when we know which rows have been sampled, i.e., when $M_s$ is provided. 
Thus, the time cost is reduced to $O(mkp)$. 
The number of rows of $M_s$, i.e., $k$, is referred to as the sketching size.
For further information on sketching preliminaries, readers are referred to \cite{martinsson2020randomized, woodruff2014sketching}.

Our work also incorporates both gradient descent and stochastic gradient descent techniques. Gradient descent is a fundamental method in optimization that iteratively updates the optimizer by moving in the opposite direction of the gradient. For example, if $f(x)$ is the function to be minimized, we update the estimated minimizer $x_{t-1}$ from the previous iteration by $x_t = x_{t-1} - \alpha \nabla f(x_{t-1})$, where $\nabla f (x_{t-1})$ is the gradient of $f$ at $x_{t-1}$ and $\alpha$ is some predetermined step size. 
Stochastic gradient descent can be seen as a stochastic approximation of gradient descent. It replaces the actual gradient $\nabla f (x_{t-1})$, calculated from the entire dataset, with an estimate calculated from a randomly selected subset of the data \cite{boyd2004convex}. In optimization problems with high dimensions, stochastic gradient descent reduces computational burden while achieving faster iterations at the cost of lower convergence rates \cite{bottou2007tradeoffs}.

%%%%%%%%%%%%%
\section{Tensor Decomposition with Unaligned Observations and $\ell_2$ Loss}\label{sec:rkhs-tensor-decomposition}
%%%%%%%%%%%%%

Before addressing the general loss function $f$ in the optimization problem \eqref{optimization_problem_generalized_sec1}, we first consider the simplified and widely studied case with the $\ell_2$ loss: $f(a, b) = (a-b)^2$. Then, \eqref{optimization_problem_generalized_sec1} reduces to
\begin{equation}\label{optimization_problem}
  \left\{\hat{a}_r, \hat{b}_r, \hat{\xi}_r\right\}_{r=1}^R = \argmin_{\substack{a_r\in \mathbb{R}^n, b_r\in \mathbb{R}^p, \xi_r\in \mathcal{H} \\ \|a_r\|=\|b_r\| =1, \|\xi_r\|_\mathcal{H} \leq \lambda \\ r = 1, \ldots, R}} %&
  \sum_{i=1}^n\sum_{j=1}^p \sum_{t\in T_i} \left(\bcX_{ij}(t) - \sum_{r=1}^R (a_r)_i\cdot (b_r)_j\cdot \xi_r(t)\right)^2. 
\end{equation} 
Here, $\|\cdot \|_{\mathcal{H}}$ denotes the RKHS norm in $\mathcal{H}$ (see Section \ref{sec:preliminary}). 
The main motivation of constraining $\|\xi_r \|_{\mathcal{H}} \leq \lambda$ in \eqref{optimization_problem} is to encourage smoothness of $\xi_r$ and to mitigate overfitting \cite{wahba1990spline}.

%%%%%%%%%%%%%
\subsection{Computation}\label{sec:computation-RKHS-tensor}
%%%%%%%%%%%%%

We employ an alternating minimization approach to solve \eqref{optimization_problem}, given its quadratic loss format. Note that $\ell_2$-ball constrained problems typically do not admit closed-form solutions, making them computationally more expensive. To improve efficiency, we update one mode ($a_r$, $b_r$, or $\xi_r$) in each iteration while keeping the other modes fixed, and this process is repeated for each mode:
\begin{align}
        \text{(Update $\{a_r\}_{r = 1}^R$)} \quad & \argmin_{A = [a_1, \ldots, a_R]} \sum_{i=1}^n\sum_{j=1}^p \sum_{t\in T_i} \left(\bcX_{ij}(t) - \sum_{r=1}^R(a_r)_i\cdot  (b_r)_j\cdot  \xi_r(t)\right)^2;\label{optimization_A}\\
        \text{(Update $\{b_r\}_{r = 1}^R$)} \quad & \argmin_{B = [b_1, \ldots, b_R]}\sum_{i=1}^n\sum_{j=1}^p \sum_{t\in T_i} \left(\bcX_{ij}(t) - \sum_{r=1}^R(a_r)_i\cdot  (b_r)_j\cdot  \xi_r(t)\right)^2; \label{optimization_B}\\
        \text{(Update $\{\xi_r\}_{r = 1}^R$)} \quad & \argmin_{\|\xi_r\|_\mathcal{H} \leq \lambda_\xi, r= 1,\ldots, R} \sum_{i=1}^n\sum_{j=1}^p \sum_{t\in T_i} \left(\bcX_{ij}(t) - \sum_{r=1}^R (a_r)_i\cdot (b_r)_j\cdot \xi_r(t)\right)^2.\label{loss_xi_2}
\end{align}
This unconstrained least-squares update followed by normalization approach allows for a closed-form solution and is significantly more efficient.

{%%%%%%%%%
\paragraph{Update Tabular Modes: $\hat{a}_r, \hat{b}_r$.}\label{sec:update-tabular-modes}
%%%%%%%%%
Denote $A = [a_1, \ldots, a_R], B = [b_1, \ldots, b_R]$ and $\Xi = [\xi_1, \ldots, \xi_R]$. 
First note that \eqref{optimization_A} can be solved with respect to each index $i$ and 
\begin{displaymath}
B \odot \Xi(T_i) = 
\begin{bmatrix}
    b_1 \otimes \xi_1(T_i), \ldots, b_R \otimes \xi_R(T_i)
\end{bmatrix}
\in \mathbb{R}^{(|T_i|p) \times R},\quad i\in [n].
\end{displaymath}
Hence, we can update $A$ row by row as
\begin{equation}\label{loss_a}
	\begin{split}
		\begin{pmatrix}
			(a_1)_i,
			\ldots,
			(a_R)_i
		\end{pmatrix}\T = \argmin_{a \in\mathbb{R}^R}\left\| \left(\bcX_{i1}(T_i)\T, \ldots, \bcX_{ip}(T_i)\T \right)^\top -  \left(B \odot \Xi(T_i)\right) a\right\|^2_2, \quad i\in [n]. 
	\end{split}
\end{equation}
Similar arguments hold for \eqref{optimization_B}, where we denote the Khatri-Rao product
\begin{displaymath}A \odot [\Xi(T_1)\T, \ldots, \Xi(T_n)\T]\T
= \begin{bmatrix}
    	(a_1)_1 \xi_1(T_1)\T &  (a_1)_2 \xi_1(T_2)\T & \ldots &  (a_1)_n \xi_1(T_n)\T\\
    	\vdots & \vdots & & \vdots\\
    	(a_R)_1 \xi_R(T_1)\T &  (a_R)_2 \xi_R(T_2)\T & \ldots &  (a_R)_n \xi_R(T_n)\T\\
    \end{bmatrix}^\top
  \end{displaymath} and then $B$ can be updated as 
\begin{equation}\notag
	\begin{split}
		\begin{pmatrix}(b_1)_j,\ldots,(b_R)_j\end{pmatrix}\T = \argmin_{b \in\mathbb{R}^R}\left\| \left(\bcX_{1j}(T_1)\T, \ldots, \bcX_{nj}(T_n)\T \right)^\top -  \left(A \odot [\Xi(T_1)\T, \ldots, \Xi(T_n)\T]\T \right) b\right\|^2_2, \quad j\in [p],
	\end{split}
\end{equation}
which can be further written into a more compact form as a matrix least square problem
\begin{equation}\label{loss_b}
    \hat{B} = \argmin_{B} \left\|X - B \left(A\odot [\Xi(T_1)\T, \ldots, \Xi(T_n)\T]\T\right)^\top \right\|_{\F}^2, 
\end{equation}
where $X$ is a matrix whose $j$th row is $\left(\bcX_{1j}(T_1)\T, \ldots, \bcX_{nj}(T_n)\T \right)$.

%%%%%%%%%%
\paragraph{Update the Functional Mode: $\hat{\xi}_r$.}\label{sec:update-functional}
%%%%%%%%%%

We first note that \eqref{loss_xi_2} is a constrained convex optimization problem. So the feasible set is bounded and closed. By Banach-Alaoglu Theorem \cite{rudin1991functional}, any bounded closed set in Hilbert space is weakly compact. Note that when fixing $a_r$ and $b_r$, the loss function is Gateaux-differentiable and convex on $\xi_r$. Hence, the minimizer exists (See Appendix Section \ref{proposition_representer_thm_proof} for detailed explanation). 
By Karush–Kuhn–Tucker (KKT) optimality condition in Hilbert Space (e.g., Theorem 5.1 in chapter 3 of \cite{ekeland1999convex}), there exists some constant $\lambda'$ such that we can replace \eqref{loss_xi_2} by
\begin{equation}\label{loss_xi_2_regularized}
    \argmin_{\xi_r\in \mathcal{H}, r= 1,\ldots, R} \sum_{i=1}^n\sum_{j=1}^p \sum_{t\in T_i} \left(\bcX_{ij}(t) - \sum_{r=1}^R (a_r)_i\cdot (b_r)_j\cdot \xi_r(t)\right)^2+\lambda'\sum_{r=1}^R \|\xi_r\|_{\mathcal{H}}^2. 
\end{equation}
By Representer Theorem \cite{scholkopf2001generalized}, the solution of \eqref{loss_xi_2_regularized} can be represented as 
\begin{equation}\label{loss_xi_representation_0}
    \hat{\xi}_r=\sum_{i = 1}^n \sum_{t \in T_i} \tilde\theta_{r, t} \mathbb{K}(\cdot, t), \quad r=1, \ldots, R,
\end{equation}
for some coefficients $\tilde\theta_{r, t}$. Here, $\mathbb{K}$ is the reproducing kernel associated with the RKHS $\mathcal{H}$. 
By collecting the repeated elements in $T_1, \ldots, T_n$, \eqref{loss_xi_representation_0} can be rewritten as 
\begin{equation}\label{loss_xi_representation}
    \hat{\xi}_r=\sum_{t \in T} \theta_{r, t} \mathbb{K}(\cdot, t), \quad r=1, \ldots, R,
\end{equation}
where $T = \cup_{i=1}^n T_i$. Compared with \eqref{loss_xi_representation_0}, the dimensionality of \eqref{loss_xi_representation} is reduced to $|T|$ from $\sum_{i = 1}^n |T_i|$.
With this parametrization, \eqref{loss_xi_2_regularized} is formatted as a computable discrete quadratic optimization problem with respect to $\theta$. 
Next, we discuss how to actually solve this optimization problem in a computing machine, i.e., express this problem in a vectorized and computable format in the closed form of vector $\theta = (\theta_{1, 1}, \theta_{1, 2}, \ldots, \theta_{1, |T|},\theta_{2, 1}, \ldots, \theta_{R, |T|})\T \in \RR^{R|T|}$.

Denote the estimated tensor as $\hat \bcX_{ij}(t) = \sum_{r=1}^R (a_r)_i\cdot (b_r)_j\cdot \xi_r(t)$. We vectorize the observed tensor and the estimated tensor respectively as 
\[
    x = \left(\bcX_{11}(T_1)\T, \ldots, \bcX_{1p}(T_1)\T, \ldots, \bcX_{i1}(T_i)\T, \ldots, \bcX_{ip}(T_i)\T , \ldots, \bcX_{n1}(T_n)\T, \ldots, \bcX_{np}(T_n)\T\right)^\top \in \RR^{|\Omega|},
\] 
\[
    \hat x = \left(\hat \bcX_{11}(T_1)\T, \ldots, \hat \bcX_{1p}(T_1)\T, \ldots, \hat \bcX_{i1}(T_i)\T, \ldots, \hat \bcX_{ip}(T_i)\T , \ldots, \hat \bcX_{n1}(T_n)\T, \ldots, \hat \bcX_{np}(T_n)\T\right)^\top \in \RR^{|\Omega|}. 
\]
We aim to express $\hat x$ in terms of $A, B$ and $\Xi(T)$. To this end, we denote the indices in the tensor of the $k$th entry of $x$ as $i_k, j_k$ and $t_k$ and define the permutation matrices $P_1 \in \RR^{|\Omega| \times n}, P_2 \in \RR^{|\Omega| \times p}$ and $P_3 \in \RR^{|\Omega| \times |T|}$. We ensure each row of $P_1, P_2, P_3$ contains exactly 1 one, with all other entries being zero. Furthermore, $(P_1 a_r)_k = (a_r)_{i_k}, (P_2 b_r)_k = (b_r)_{j_k}$ and $(P_3 \xi_r(T))_k = \xi_r(t_k)$. Then, we have $[(P_1 A) * (P_2 B) * (P_3 \Xi(T))]_{k,r} = (a_r)_{i_k}(b_r)_{j_k}\xi_r(t_k)$ and hence, row sum of the $k$th row in $[(P_1 A) * (P_2 B) * (P_3 \Xi(T))]$ is the $k$th entry of $\hat x$. Thus, we can rewrite \eqref{loss_xi_2_regularized} as
\begin{equation}\label{loss_xi_2_regularized_alt}
  \argmin_{\Xi}\left\| x - [(P_1 A) * (P_2 B) * (P_3 \Xi(T))] \boldsymbol{1}_R\right\|^2+\lambda'\sum_{r=1}^R \|\xi_r\|_{\mathcal{H}}^2,
\end{equation} 
where $ \boldsymbol{1}_R \in \RR^R$ is a vector with 1 in all entries. Plugging the representation \eqref{loss_xi_representation} into \eqref{loss_xi_2_regularized_alt}, we can update the functional mode by solving
\begin{equation}\label{eq:loss-function-xi}
     \hat \theta = \argmin_{\theta \in \RR^{R|T|}}\left\| x - D\theta\right\|^2+\lambda'\theta\T\tilde{K} \theta,
\end{equation}
where $\tilde{K} \in \RR^{R|T|\times R|T|}$ is a blocked diagonal matrix with $R$ same blocks in its diagonal and each of the block is $\mathbb{K}(T, T) \in \RR^{|T|\times |T|}$, and $D = [D_1, \ldots, D_R]$ where $D_r = (P_1 a_r) * (P_2 b_r) * (P_3 \mathbb{K}(T, T))$ and the Hadamard product between a vector and a matrix is defined earlier in Section \ref{sec:preliminary}. Hence, we have
\[
\hat{\xi}_r=\sum_{s = 1}^{|T|} \hat \theta_{r, s} \mathbb{K}(\cdot, t_s), \quad r=1, \ldots, R, 
\]
where $\hat{\theta} = (\hat\theta_{1, 1}, \hat\theta_{1, 2}, \ldots, \hat\theta_{1, |T|},\hat\theta_{2, 1}, \ldots, \hat\theta_{R, |T|})\T$ and $t_s$ is the $s$th element in $T$.

The overall procedure of our proposed tensor decomposition with unaligned observations is summarized to \cref{algorithm_RKHS}.
\begin{algorithm}
	\caption{Tensor Decomposition with Unaligned Observations via RKHS (RKHS-TD)}
	\label{algorithm_RKHS}
        \algrenewcommand\algorithmicensure{\textbf{Output:}}
        \algrenewcommand\algorithmicrequire{\textbf{Input:}}
	\begin{algorithmic}[1]
        \Require{Observed functional tensor $\bcX_{ij}(t)$; Penalty coefficient  $\lambda'$; Target rank $R$; Maximum iterations $m_{\max}$; Initialization $A, B, \theta$}
	\Ensure $A, B, \theta$ and $\hat \bcX_{ij}(t)$
	\For{$t$ in  $1, \ldots, m_{\max}$}
        \State {Update $A$ by \eqref{loss_a} and normalize;}
        \State {Update $B$ by \eqref{loss_b} and normalize;}
        \State {Update $\theta$ by \eqref{eq:loss-function-xi};}
        \EndFor\\
        \Return $A, B, \theta$ and $\hat \bcX_{ij}(t)$
	\end{algorithmic}
\end{algorithm}

%%%%%%%%%%%%
\paragraph{Time Complexity.}\label{sec:complexity}
%%%%%%%%%%%%

In each iteration of \cref{algorithm_RKHS}, the computation of $\Xi(T_i)$ for all $i$ in \eqref{loss_a} and \eqref{loss_b} takes $R|T|\sum_{i= 1}^n|T_i|$ floating-point operations (flops). 
In \eqref{loss_a}, computing the column-wise Khatri-Rao product of $B\in \RR^{p\times R}$ and $\Xi(T_i) \in \RR^{|T_i|\times R}$ takes $O(pR|T_i|)$ flops, and solving $a$ via Cholesky decomposition takes $O(p|T_i|R^2)$ flops. 
Thus, the total cost to update $A$ is $O(pR^2\sum_{i = 1}^n|T_i|)$ flops. 

In \eqref{loss_b}, calculating $A\odot\left[ \Xi(T_1)\T, \ldots,  \Xi(T_n)\T \right]\T$ requires $O(pR\sum_{i = 1}^n|T_i|)$ and solving $B$ requires $O(pR^2\sum_{i = 1}^n|T_i|)$ flops, which is the overall cost to update $B$. 

For the functional mode, computing $D_r$ for each $r$ takes $O(p|T|\sum_{i= 1}^n|T_i|)$, and hence obtaining $D$ costs $O(pR|T|\sum_{i= 1}^n|T_i|)$. To solve \eqref{eq:loss-function-xi} by its normal equation, we need to calculate $D\T D$, which takes $O(pR^2|T|^2\sum_{i= 1}^n|T_i|)$ flops. Finally, solving the normal equation takes $O(R^2|T|^2(R|T| + p\sum_{i= 1}^n|T_i|))$ flops.

Therefore, assuming $p\sum_{i= 1}^n|T_i| \geq R |T|$, each iteration in RKHS-TD (\cref{algorithm_RKHS}) requires $O(pR^2|T|^2\sum_{i= 1}^n|T_i|)$ flops. }

    It is worth noting that when the observations are aligned, i.e., $T_1 = \cdots = T_n = T$, the computational complexity can be significantly reduced, as $|T| = |T_i|$. In contrast, when the observations are more unaligned, $|T|$ becomes larger, leading to increased computational cost. For further details, see \cite{larsen2024tensor}.

%%%%%%%%%%%%%%
\subsection{Fast Computation via Sketchings}\label{sec:sketching}
%%%%%%%%%%%%%%
As highlighted in \cref{sec:intro}, irregularly observed time points in the unaligned mode can significantly increase the computational time required by the method. For instance, consider the scenario described in \cref{example_clinical_trail}. If all patients attend their tests as scheduled, we have $T_i = T_j$ with $|T_i| = 5$. With only five observed time points, the optimization problem in \eqref{loss_xi_2_regularized} during each iteration of \cref{algorithm_RKHS} is limited to a five-dimensional space (assuming the target rank $R = 1$). However, if the actual day for each patient to attend the $i$th test varies between $30 \times (i - 1) - 5$ and $30 \times (i - 1) + 5$, the dimensionality of \eqref{loss_xi_2_regularized} expands to 55. In the general case, the dimension of $\theta$ is $R|T|$ where $T = \bigcup_{i = 1}^n T_i$ is the set of all observed time points and often has a high cardinality. This substantial increase in dimension significantly raises the computation time, as \eqref{eq:loss-function-xi} must be repeatedly solved through iterations. 

To tackle this issue, we propose to apply the sketching technique to \cref{algorithm_RKHS}. 
Specifically when updating $\hat{a}$, we generate sketching matrices $S_i \in \RR^{k_a \times p|T_i|}$ for each $i$, where $k_a$ is some pre-specified sketching size. Then we apply sketching using $S_i$ to \eqref{loss_a}: 
\[
\begin{split}
		\begin{pmatrix}
			(a_1)_i\\
			\vdots\\
			(a_R)_i
		\end{pmatrix} = \argmin_{a \in\mathbb{R}^R}\left\| S_i \left(\bcX_{i1}(T_i)\T, \ldots, \bcX_{ip}(T_i)\T \right)^\top -  S_i \left(B \odot \Xi(T_i)\right) a\right\|^2_2, \quad i\in [n].
	\end{split}
\]
A similar sketching approach can also be applied when updating $b$. 
To update $\xi_r$, we randomly generate a sketching matrix $S \in \RR^{k_\xi \times p\sum_{i = 1}^n|T_i|}$, where $k_\xi$ is some pre-specified sketching size. Then we address the following sketched version of $\eqref{eq:loss-function-xi}$:
\begin{equation}\label{eq:loss-function-xi-sketched}
	\argmin_{\theta \in \RR^{R|T|}}\left\| Sx - SD\theta\right\|^2+\lambda'\theta\T\tilde{K} \theta,. 
\end{equation}
Instead of directly generating $S$, individually computing each matrix in the product $SD$, and then multiplying them together, we further propose the following sampling procedure to expedite this process, taking advantage of the structure of each matrix. Given an integer $|\hat T_i|$ for each $i$, we uniformly sample $t_1, \dots, t_{|\hat T_i|}$ i.i.d. from $T_i$ and use them to form  set $\hat T_i$. We also uniformly sample subset $[\hat N] \subseteq[n]$ and $[\hat J] \subseteq[p]$. Now, we use sampled observations $\bcX_{ij}(t), i \in \hat N, j \in \hat J$ and $t_i \in \hat T_i$ to form the new unaligned tensor, use this new tensor to calculate $x$ and $D$ in \eqref{eq:loss-function-xi-sketched}, and finally use them in each iteration of \cref{algorithm_RKHS}. The detailed description of the correspondence between the sketching matrix and the proposed sampling procedure and the resulting sketching algorithm (S-RKHS-TD, \cref{algorithm_RKHS_sketched}) are provided in \cref{sec:supplement_sketching}. We provide the time complexity comparison in \cref{table_complexity_RKHS}. Notably, as we usually have $\sum_{i = 1}^n|T_i| \gg |T|$ in practice, the sketched algorithm would be faster.

\begin{table}[!ht]
    \centering
    \begin{tabular}{cc}\toprule
        \textbf{Algorithm} & \textbf{Time Complexity per Iteration} \\ \hline
        RKHS-TD ({\cref{algorithm_RKHS}}) &  $O(pR^2|T|^2\sum_{i = 1}^n|T_i|)$ \\ 
        S-RKHS-TD ({\cref{algorithm_RKHS_sketched}}) &  $O(R^2|T|^2(R|T| + |\hat J| \sum_{i\in \hat{N}}|\hat{T}_i|))$\\
        \bottomrule
    \end{tabular}
    \caption{Comparison of time complexity based on input functional tensor size ($p\times \sum_{i = 1}^n |T_i|$) and target rank ($R$). In the S-RKHS-TD method, we use sampled indices $\hat{N} \in [n]$, $\hat{J} \in [p]$, and $\hat{T}_i \in T_i$ for $i\in \hat{N}$.}
    \label{table_complexity_RKHS}
\end{table}

%%%%%%%%%%%%
\section{Tensor Decomposition with Unaligned Observations and General Loss}\label{sec:formulation-generalized}
%%%%%%%%%%%%
This section studies the tensor decomposition with unaligned observations and general loss function as described in \eqref{optimization_problem_generalized_sec1}. This approach encompasses a broader setting of tensor decomposition, where observations follow a more general class of distributions. We first restate \eqref{optimization_problem_generalized_sec1}: 
\begin{equation}\label{optimization_problem_generalized}
    \{\hat{a}_r, \hat{b}_r, \hat{\xi}_r\}_{r =1}^R = \argmin_{\substack{(a_r, b_r, \xi_r) \in \Phi \\ {r = 1,\ldots,R}}} \frac{1}{|\Omega|}\sum_{i=1}^n\sum_{j=1}^p \sum_{t\in T_i} f\left(\sum_{r=1}^R (a_r)_i\cdot (b_r)_j\cdot \xi_r(t), \bcX_{ij}(t)\right).
\end{equation}
Here, $\Phi$ represents a feasible set. To ensure the smoothness of $\xi_r$ and prevent overfitting, and most importantly, we propose constraining the RKHS norm of $\xi_r$ after properly normalizing $a_r$ and $b_r$. In other words, we enforce $\|a_r\| \|b_r\| \|\xi_r\|_{\mathcal{H}} \leq \lambda$ for constant $\lambda$. Consequently, $\Phi = (\mathbb{R}^n \times \mathbb{R}^p \times \mathcal{H})\cap \{\max_r\|a_r\|\|b_r\|\|\xi_r\|_{\mathcal{H}} \leq \lambda\} \cap \Phi_0$, where $\Phi_0$ represents other possible constraints introduced by the specific loss function $f$. We refer to \eqref{optimization_problem_generalized} as the generalized tensor decomposition with unaligned observations.

%%%%%%%%%%%
\subsection{Computation via Gradient Descent}\label{sec:computation-grkhs}
%%%%%%%%%%%

A generalized form of the Representer Theorem \cite{scholkopf2001generalized} provides a fundamental representation of the solution to \eqref{optimization_problem_generalized}, listed in the following proposition.
\begin{proposition} \label{proposition_representer_thm}
    If $f(x,y)$ is a convex and Gateaux-differentiable function of $x$ for given $y$, then the optimization problem \eqref{optimization_problem_generalized} admits a solution in which $\hat\xi_r$ can be represented as $\hat{\xi}_r=\sum_{s \in \cup_{i=1}^n T_i} \theta_{r, s} \mathbb{K}(\cdot, s)$ for $\theta=(\theta_{r, s})\in \RR^{\sum_{i=1}^n |T_i|}$, where $\mathcal{H}$ is a RKHS with kernel $\mathbb{K}(\cdot, \cdot)$. 
\end{proposition}

Denote the loss function
\[
F(a_r, b_r, \xi_r) = \frac{1}{|\Omega|}\sum_{i=1}^n\sum_{j=1}^p \sum_{t\in T_i} f\left(\sum_{r=1}^R (a_r)_i\cdot (b_r)_j\cdot \xi_r(t), \bcX_{ij}(t)\right). 
\]
Plugging in $\xi_r = \sum_{s \in \cup_{i=1}^n T_i} \theta_{r, s}K(t,s)$ to $F$, we obtain 
\be\label{eq_parameterized_generalized_loss}
\tilde F = \frac{1}{|\Omega|}\sum_{i=1}^n\sum_{j=1}^p \sum_{t\in T_i} f\left(\sum_{r=1}^R (a_r)_i\cdot (b_r)_j\cdot \sum_{s \in T} \theta_{r, s} \mathbb{K}(t, s), \bcX_{ij}(t)\right). 
\ee
Then solving \eqref{optimization_problem_generalized} is essentially minimizing \eqref{eq_parameterized_generalized_loss}. Naturally, we can apply gradient descent to minimize \eqref{eq_parameterized_generalized_loss}. Utilizing the notation introduced in \cref{sec:computation-RKHS-tensor}, we can express the gradient as:
    \begin{align}
	\begin{pmatrix}
		\frac{\partial \tilde F}{\partial \left({a}_1\right)_k}\\
		\vdots\\
		\frac{\partial \tilde F}{\partial \left({a}_R\right)_k}
	\end{pmatrix} 
	= \frac{1}{|\Omega|}(B \odot \Xi(T_i))\T \begin{pmatrix}
		\frac{\partial f(\hat{\bcX}_{k,1}(T_k), \bcX_{k,1}(T_k))}{\partial \hat{\bcX}_{k,1}(T_k)}\\
		\vdots\\
		\frac{\partial f(\hat{\bcX}_{k,p}(T_k), \bcX_{k,p}(T_k))}{\partial \hat{\bcX}_{k,p}(T_k)} \label{eq_gradient_a}
	\end{pmatrix}; 
    \end{align}	
    \begin{equation}
	\frac{\partial \tilde F}{\partial B} = 
	\begin{bmatrix}
		\left(\frac{\partial f(\hat{\bcX}_{1,1}(T_1), \bcX_{1, 1}(T_1))}{\partial \hat{\bcX}_{1,1}(T_1)}\right)\T &\ldots& \left(\frac{\partial f(\hat{\bcX}_{n,1}(T_n), \bcX_{n, 1}(T_n))}{\partial \hat{\bcX}_{n,1}(T_n)}\right)\T\\
		\vdots && \vdots \\
		\left(\frac{\partial f(\hat{\bcX}_{1,p}(T_1), \bcX_{1, p}(T_1))}{\partial \hat{\bcX}_{1,p}(T_1)}\right)\T &\ldots& \left(\frac{\partial f(\hat{\bcX}_{n,p}(T_n), \bcX_{n, p}(T_n))}{\partial \hat{\bcX}_{n,p}(T_n)}\right)\T\\
	\end{bmatrix} \left(A \odot \begin{bmatrix}
\Xi(T_1)\\ 
\vdots\\  
\Xi(T_n)
\end{bmatrix}\right); \label{eq_gradient_b}
    \end{equation}
\begin{equation}\label{eq_gradient_theta}
	\frac{\partial \tilde F}{\partial {\theta}} = 
	\frac{1}{|\Omega|} D\T
	\left(\frac{\partial f(\hat{\bcX}_{i,j}(t), \bcX_{i,j}(t))}{\partial \hat{\bcX}_{k,j}(t)}\right)_{j,i,t\in T_i}.
    \end{equation}
Here $\frac{\partial \tilde F}{\partial B}$ denotes the $p$-by-$R$ matrix of partial gradients with $[\frac{\partial \tilde F}{\partial B}]_{j,r} = \frac{\partial \tilde F}{\partial ({b}_r)_j}$, and $\frac{\partial \tilde F}{\partial {\theta}}$ is the $R|T|$-dimensional vector of partial gradients: $\frac{\partial \tilde F}{\partial {\theta}} = (\frac{\partial \tilde F}{\partial {\theta_{1, 1}}}, \frac{\partial \tilde F}{\partial {\theta_{1, 2}}}, \ldots, \frac{\partial \tilde F}{\partial {\theta_{R, |T|}}})\T \in \RR^{R|T|}$. We also define matrix $\frac{\partial \tilde F}{\partial A} \in \RR^{n\times R}$ as $[\frac{\partial \tilde F}{\partial A}]_{i,r} = \frac{\partial \tilde F}{\partial ({a}_r)_i}$. 

Then, we can perform gradient descent using \eqref{eq_gradient_a}, \eqref{eq_gradient_b}, and \eqref{eq_gradient_theta}. To take the feasible set $\Phi \subseteq \{\max_r\|a_r\|\|b_r\|\|\xi_r\|_{\mathcal{H}} \leq \lambda\}$ into account, we first scale $a_r, b_r$ and $\xi_r$ by $a_r' = a_r\frac{\lambda^{1/3}}{\|a_r\|}$, $b_r' = b_r\frac{\lambda^{1/3}}{\|b_r\|}$ and $\xi_r' = \xi_r\frac{\lambda^{1/3}}{\|\xi_r\|_{\mathcal{H}}}$ if $\|a_r\|\|b_r\|\|\xi_r\|_{\mathcal{H}} > \lambda$; and then project $a_r', b_r'$ and $\xi_r'$ to $\Phi$ if $\Phi \subsetneq \{\max_r \|a_r\|\|b_r\|\|\xi_r\|_{\mathcal{H}} \leq \lambda\}$ after each update. This procedure is summarised to \cref{algorithm_PGD}. 
\begin{algorithm}[h]
    \caption{Gradient Descent with Scaling and Projection for Generalized Functional Tensor Decomposition via RKHS (GRKHS-TD)}
	\label{algorithm_PGD}
	\begin{algorithmic}[1]
        \algrenewcommand\algorithmicensure{\textbf{Output:}}
        \algrenewcommand\algorithmicrequire{\textbf{Input:}}
		\Require {Observed functional tensor $\bcX_{ij}(t)$; pairwise loss function $f$; feasible set $\Phi$; learning rate $\alpha$; target rank $R$; maximum iterations $m_{\max}$;
        initialization $A, B, \theta$}
		\Ensure $A, B, \theta$ and $\hat \bcX_{ij}(t)$
		\For{$t$ in  $1, \ldots, m_{\max}$}
        \State{Calculate $\frac{\partial \tilde F}{\partial {A}}$, $\frac{\partial \tilde F}{\partial {B}}$, and $\frac{\partial \tilde F}{\partial {\theta}}$ by \eqref{eq_gradient_a}, \eqref{eq_gradient_b}, and \eqref{eq_gradient_theta}, respectively. \label{a_4_grad}}
		\State{Let ${A} \leftarrow A - \alpha \frac{\partial \tilde F}{\partial A}$, ${B} \leftarrow B - \alpha \frac{\partial \tilde F}{\partial B}$ and ${\theta} \leftarrow \theta - \alpha \frac{\partial \tilde F}{\partial \theta}$;}
            \If {$\|a_r\|\|b_r\|\|\xi_r\|_{\mathcal{H}} > C$ for any $r$}
		\State {$a_r \leftarrow a_r\frac{C^{1/3}}{\|a_r\|}$, $b_r \leftarrow b_r\frac{C^{1/3}}{\|b_r\|}$ and $\xi_r \leftarrow \xi_r\frac{C^{1/3}}{\|\xi_r\|_{\mathcal{H}}}$;\label{a_4_normalization}}
		\EndIf
		\State{Project $A$, $B$ and $\theta$ to feasible set $\Phi$;}
		\EndFor\\
		\Return $A, B, \theta$ and $\hat \bcX_{ij}(t)$
	\end{algorithmic}
\end{algorithm}

\paragraph{Time Complexity.} In \cref{algorithm_PGD}, \cref{a_4_grad} involves calculating the partial gradient of $f\left(\hat\bcX_{ij}(t), \bcX_{ij}(t)\right)$ for $i \in [n]$, $j =[p]$, and $t \in T_i$. This computation requires $O(p\sum_{i = 1}^n|T_i|)$ flops. The computation of $\frac{\partial \tilde F}{\partial A}$, $\frac{\partial \tilde F}{\partial B}$, and $\frac{\partial \tilde F}{\partial {\theta}}$ using \eqref{eq_gradient_a}, \eqref{eq_gradient_b}, and \eqref{eq_gradient_theta} respectively, requires $O(p|T|R\sum_{i = 1}^n|T_i|)$ flops. This computational load characterizes each iteration of \cref{algorithm_PGD}, assuming that the computation involving the projection to the feasible set $\Phi$ does not dominate the overall process.

\paragraph{Normalization}
In \cref{algorithm_PGD}, we normalize the product of the factors $a$, $b$, and $\xi$, followed by a balanced reweighting as described in \cref{a_4_normalization}. A natural question is why we do not normalize each component individually. Empirically, we observed that separate normalization can lead to instability: it often only rescales the dominant factor, allowing the smaller factor to remain small or even shrink further over iterations. In contrast, our joint normalization approach balances the magnitudes of all factors simultaneously, which leads to improved numerical stability and better empirical performance.

\subsection{Fast Computation via Stochastic Gradient Descent} \label{section_fast_GRKHS_computation}

We further propose the stochastic gradient descent approach to accelerate the computation of the optimization problem \eqref{optimization_problem_generalized}. Specifically in each iteration, we uniformly sample subsets $\hat{N} \subseteq [n], \hat{J} \subseteq [p]$ and $\hat{T_i} \subseteq T_i$ for all $i \in [n]$. 
Then evaluate the sketched gradient by replacing $\left(\frac{\partial f(\hat{\bcX}_{i,j}(t), \bcX_{i,j}(t))}{\partial \hat{\bcX}_{k,j}(t)}\right)$ by zero for $i\notin \hat N$, $j \notin \hat J$ or $t\notin \hat T_i$ in \eqref{eq_gradient_a}, \eqref{eq_gradient_b}, and \eqref{eq_gradient_theta}. Similar to the S-RKHS-TD discussed in \cref{sec:sketching}, the sketched gradient $\frac{\partial \tilde F}{\partial {\theta}}$ can be more efficiently calculated by:
\begin{equation}\label{eq_gradient_theta_sketched}
    \frac{\partial \tilde F}{\partial {\theta}} = 
	\frac{1}{|\{(i,j,t): i\in \hat N, j \in \hat J, t\in \hat T_i\}|} (S D)\T
	\left(\frac{\partial f(\hat{\bcX}_{i,j}(t), \bcX_{i,j}(t))}{\partial \hat{\bcX}_{k,j}(t)}\right)_{i\in \hat N, j \in \hat J, t\in \hat T_i}, 
\end{equation}
where $S$ is the sketching matrix discussed in \cref{sec:sketching}. The overall matrix product $SD$ above can be calculated by \cref{algorithm_sketched_coefficient} in the \cref{sec:supplement_sketching}. We summarize the overall procedure to \cref{algorithm_PSGD} and refer to it briefly as S-GRHKS-TD.  
\begin{algorithm}
    \caption{Stochastic Gradient Descent with Scaling and Projection for Generalized Tensor Decomposition with Unaligned Observations via RKHS (S-GRKHS-TD)}
	\label{algorithm_PSGD}
	\begin{algorithmic}[1]
        \algrenewcommand\algorithmicensure{\textbf{Output:}}
        \algrenewcommand\algorithmicrequire{\textbf{Input:}}
		\Require {Observed functional tensor $\bcX_{ij}(t)$; sample size $\hat{N}$, $\hat{J}$ and $\hat{T_i}$ for $i\in \hat{N}$; pairwise loss function $f$; feasible set $\Phi$; learning rate $\alpha$; target rank $R$; maximum iterations $m_{\max}$; initialization $A, B, \theta$}
		\Ensure $A, B, \theta$ and $\hat \bcX_{ij}(t)$
		\For{$t$ in  $1, \ldots, m_{\max}$}
		\State{Sample subsets $\hat{N} \subseteq{[n]}, \hat{J} \subseteq [p]$ and $\hat{T_i} \subseteq T_i$ for $i\in \hat{N} $;}
		\For{$i$ in  $1, \ldots, n$; $j$ in  $1, \ldots, p$; $t$ in  $1, \ldots, |T_i|$}
		\If {$i \in \hat{N}$ and $j \in \hat{J}$ and $t \in \hat{T_i}$}
		\State {Calculate $\frac{\partial f(\hat{\bcX}_{i,j}(t), \bcX_{i,j}(t))}{\partial \hat{\bcX}_{k,j}(t)}$;}
		\Else 
		\State {$\frac{\partial f(\hat{\bcX}_{i,j}(t), \bcX_{i,j}(t))}{\partial \hat{\bcX}_{k,j}(t)} \leftarrow 0$;}
		\EndIf
		\EndFor
		\State{Calculate $\frac{\partial F}{\partial A}$, $\frac{\partial F}{\partial B}$ and $\frac{\partial F}{\partial {\theta}}$ by \eqref{eq_gradient_a}, \eqref{eq_gradient_b} and \eqref{eq_gradient_theta_sketched}, where $\frac{\partial f(\hat{\bcX}_{i,j}(t), \bcX_{i,j}(t))}{\partial \hat{\bcX}_{k,j}(t)}$ is calculated in the previous step, and $S D$ in \eqref{eq_gradient_theta_sketched} is calculated by \cref{algorithm_sketched_coefficient};}
		\State{Let ${A} \leftarrow A - \alpha \frac{p\sum_{i = 1}^{n}|T_i|}{|\hat J| \sum_{i \in \hat N} |\hat{T}_i|} \frac{\partial F}{\partial A}$, ${B} \leftarrow B - \alpha \frac{p\sum_{i = 1}^{n}|T_i|}{|\hat J| \sum_{i \in \hat N} |\hat{T}_i|} \frac{\partial F}{\partial B}$ and ${\theta} \leftarrow \theta - \alpha \frac{p\sum_{i = 1}^{n}|T_i|}{|\hat J| \sum_{i \in \hat N} |\hat{T}_i|} \frac{\partial F}{\partial \theta}$;}
            \If {$\|a_r\|\|b_r\|\|\xi_r\|_{\mathcal{H}} > C$ for any $r$}
		\State {$a_r \leftarrow a_r\frac{C^{1/3}}{\|a_r\|}$, $b_r \leftarrow b_r\frac{C^{1/3}}{\|b_r\|}$ and $\xi_r \leftarrow \xi_r\frac{C^{1/3}}{\|\xi_r\|_{\mathcal{H}}}$;}
		\EndIf
		\State{Project $A$, $B$ and $\theta$ to feasible set $\Phi$;}
		\EndFor\\
		\Return $A, B, \theta$ and $\hat \bcX_{ij}(t)$
	\end{algorithmic}
\end{algorithm}

%%%%%%%%%%%%%%
\paragraph{Time Complexity.}
%%%%%%%%%%%%%%

In \cref{algorithm_PSGD}, it takes $O(R |\hat J| (\sum_{i\in \hat{N}}|\hat{T}_i|)|T|)$ flops to calculate \eqref{eq_gradient_theta_sketched} by \cref{algorithm_sketched_coefficient}, which is also the cost for each iteration of \cref{algorithm_PSGD}. The comparison between \cref{algorithm_PGD,algorithm_PSGD} is summarised in \cref{table_complexity_GRKHS}. 
\begin{table}[!ht]
    \centering
    \begin{tabular}{cc}\toprule
        \textbf{Algorithm} & \textbf{Time Complexity per Iteration} \\ \hline
        GRKHS-TD ({\cref{algorithm_PGD}}) & $O(pR|T|\sum_{i = 1}^n|T_i|)$ \\  
        S-GRKHS-TD ({\cref{algorithm_PSGD}})& $O(R |\hat J| |T|\sum_{i\in \hat{N}}|\hat{T}_i|)$ \\ 
        \bottomrule
    \end{tabular}
    \caption{Time complexity comparison with input functional tensor size $p\times \sum_{i = 1}^n |T_i|$ and target rank $R$. 
    Let $\hat{N} \in{[n]}, \hat{J} \in [p]$, and $\hat{T_i} \in T_i$ for $i\in \hat{N}$, which represent the sampled indices used in S-GRKHS-TD. }
    \label{table_complexity_GRKHS}
\end{table}

\subsection{Examples}

It is noteworthy that the loss function in the optimization problem \eqref{optimization_problem_generalized} can be interpreted as the negative log-likelihood function, thereby transforming the solution to \eqref{optimization_problem_generalized} into the maximum likelihood estimate (MLE) from a statistical perspective. For example: 
\begin{Example}[Gaussian tensor decomposition with unaligned observations]\label{example_ls}
    If we set the pairwise loss function in the optimization problem \eqref{optimization_problem_generalized} to be the least squares loss, i.e., $f(x,y) = (x-y)^2$, solving \eqref{optimization_problem_generalized} is equivalent to maximizing the likelihood function over the parameters $a_r$, $b_r$, and $\xi_r$ when we assume the observations follow a Gaussian distribution: $\bcX_{ij}(t) \overset{\text{ind.}}{\sim} \operatorname{Normal}({\mathbf {\Lambda}}_{ij}(t), \sigma^2)$, 
    where $\mathbf{\Lambda}_{ij}(t) = \mathbb{E}\bcX_{ij}(t)$ satisfies $\mathbf{\Lambda}_{ij}(t) = \sum_{r=1}^R (a_r)_i\cdot (b_r)_j\cdot \xi_r(t)$.
    In this context, the feasible set is defined as $\Phi = (\mathbb{R}^n \times \mathbb{R}^p \times \mathcal{H}) \cap \{\max_r\|a_r\|\|b_r\|\|\xi_r\|_{\mathcal{H}} \leq \lambda\}$. We term this problem the Gaussian tensor decomposition with unaligned observations, which corresponds to the regular tensor decomposition with unaligned observations \eqref{optimization_problem} discussed in \cref{sec:rkhs-tensor-decomposition}.  
    Since $\{(a_r, b_r, \xi_r)\}_{r = 1}^R$ and $\{(\lambda_1a_r, \lambda_2b_r, (\lambda_1\lambda_2)^{-1}\xi_r)\}_{r = 1}^R$ yield the same solution to the optimization problem \eqref{optimization_problem} for any $\lambda_1, \lambda_2$, the constraint $\{\|a_r\| = \|b_r\|=1, \|\xi_r\|_{\mathcal{H}} \leq \lambda; r= 1\ldots, R\}$ in optimization problem \eqref{optimization_problem} is equivalent to $\{\|a_r\| \|b_r\| \|\xi_r\|_{\mathcal{H}} \leq \lambda; r= 1\ldots, R\}$. Therefore, \eqref{optimization_problem} can be considered as a special case of \eqref{optimization_problem_generalized} under the setting of Gaussian tensor decomposition with unaligned observations.
\end{Example}

\begin{Example}[Bernoulli tensor decomposition with unaligned observations]\label{example_Bernoulli}
    For binary data, we propose to use the Bernoulli loss function with logic link $f(x,y) = \log(1+\exp y)- x\times y$. Assume the data are Bernoulli distributed $\bcX_{ij}(t) \overset{\text{ind.}}{\sim} \text{Bernoulli}\left(\mathbfcal{P}_{ij}(t)\right)$. Then, by the logic link function $\mathbfcal{P}_{ij}(t) = {\exp\mathbf{\Lambda}_{ij}(t)}/{(1+\exp \mathbf{\Lambda}_{ij}(t))}, $
    we have 
    \begin{align}
        \log \operatorname{-likelihood }
        &=\sum_{i,j,t\in T_i} \bcX_{ij}(t) \log \mathbfcal{P}_{ij}(t)+\sum_{i,j,t\in T_i} (1-\bcX_{ij}(t)) \log \left(1-\mathbfcal{P}_{ij}(t)\right)\notag\\
        &=-\sum_{i,j,t\in T_i} \log(1+\exp \mathbf{\Lambda}_{ij}(t))+\sum_{i,j,t\in T_i} \bcX_{ij}(t) \mathbf{\Lambda}_{ij}(t).\notag
    \end{align}
    Thus, if we further assume $\mathbf{\Lambda}_{ij}(t)$ has a low-rank structure, i.e., $\mathbf{\Lambda}_{ij}(t) = \sum_{r=1}^R (a_r)_i\cdot (b_r)_j\cdot \xi_r(t)$, optimizing \eqref{optimization_problem_generalized} is equivalent to maximizing the log-likelihood function. The feasible set is defined as follows in this context: $\Phi = (\mathbb{R}^n \times \mathbb{R}^p \times \mathcal{H}) \cap \{\max_r\|a_r\|\|b_r\|\|\xi_r\|_{\mathcal{H}} \leq \lambda\}$.
\end{Example}

\begin{Example}[Poisson tensor decomposition with unaligned observations]\label{example_poisson}
  The Poisson loss $f(x, y) = y - x\cdot \log(y)$ can be used in the framework of \eqref{optimization_problem_generalized} to deal with counting data, and to ensure non-negativity of the parameters,
  we introduce the set $\Phi = \{a_r \geq 0, b_r \geq 0, \theta_{r,s} \geq 0\} \cap \{ \max_r\|a_r\|\|b_r\|\|\xi_r\|_{\mathcal{H}} \leq C\}$ and use the radial kernel $\mathbb{K}(s,t) = \exp(-|s-t|^2)$. Note that, as we are projecting estimated $A, B$ and $\theta$ onto $\{a_r \geq 0, b_r \geq 0, \theta_{r,s} \geq 0\}$, $\hat \bcX_{ij}(t)$ can frequently be 0, particularly when the input data tensor is sparse, but $f(x,0)$ is not well-defined. Thus, in practice, we replace $f$ with $f_\delta(x,y) = y+\delta-x \log(y+\delta)$ for some small $\delta>0$ to prevent numerical issues. To further improve numerical performance, we can apply the gradient clipping: in each iteration of \cref{algorithm_PSGD}, if $\|\frac{\partial F}{\partial A}\| > c$ for some constant $c>0$, we let $\widehat{\frac{\partial F}{\partial A}} = c \frac{\partial F}{\partial A} / \|\frac{\partial F}{\partial A}\|$ and update $A$ by $A = A - \alpha \frac{p\sum_{i = 1}^{n}|T_i|}{|\hat J| \sum_{i \in \hat N} |\hat{T}_i|} \widehat{\frac{\partial F}{\partial A}}$. And we do the same thing for $B$ and $\theta$. In this case, optimizing the loss function in \eqref{optimization_problem_generalized} is equivalent to the maximization of the likelihood with respect to the parameters $a_r$, $b_r$, and $\xi_r$ under the assumption that the data are Poisson distributed $\bcX_{ij}(t) \overset{\text{ind.}}{\sim} \operatorname{Poisson}(\mathbf{\Lambda}_{ij}(t)+\delta)$ and that the shifted expectation has a low-rank structure $\mathbf{\Lambda}_{ij}(t) = \sum_{r=1}^R (a_r)_i\cdot (b_r)_j\cdot \xi_r(t)$. 
\end{Example}
Additionally, for diverse objectives, a variety of other loss functions can be employed, including:
\begin{Example}[Non-negative tensor decomposition with unaligned observations and Beta divergence loss]\label{example_beta_d}
Beta divergence loss is extensively utilized in the non-negative matrix or tensor decomposition, especially for modeling proportions \cite{cichocki2007non, fevotte2011algorithms}. This loss function is applicable within our framework. We define the pairwise Beta divergence loss as follows:
\begin{equation}\label{eq:beta-divergence}
f(x,y) = \frac{1}{\beta(\beta-1)}\left(x^\beta+(\beta-1) y^\beta-\beta x y^{\beta-1}\right), \quad \beta \in \mathbb{R} \backslash\{0,1\}.
\end{equation}
To maintain non-negativity in the parameters, we adopt the set $\Phi = \{a_r \geq 0, b_r \geq 0, \theta_{r,s} \geq 0\} \cap \{ \max_r\|a_r\|\|b_r\|\|\xi_r\|_{\mathcal{H}} \leq \lambda\}$ the radial kernel, similar to the approach in \cref{example_poisson}. However, when $\beta < 1$, the loss function becomes undefined for $f(x,0)$. To circumvent this, we modify $f$ to $f_\delta(x,y) = f(x+\delta,y+\delta)$, using a small positive $\delta$, thus avoiding numerical issues. It is important to note that for $f_\delta(\bcX_{ij}(t),\hat \bcX_{ij}(t))$ with $\beta < 1$, if $\hat \bcX_{ij}(t) = 0$ while $\bcX_{ij}(t) \neq 0$, the loss becomes $f(\bcX_{ij}(t)+\delta,\delta)$, which approaches infinity as $\delta$ tends to zero. 
Thus, a small $\delta$ and a less-than-one $\beta$ would benefit the scenario of sparse data, which would result in a large loss when we mistakenly estimate those nonzero entries as zero. On the contrary, a naive estimation of $\hat \bcX_{ij}(t) \equiv 0$ typically results in a low loss with other common loss functions like the least squares loss or the Beta divergence loss with $\beta > 1$. 
This scenario is further explored in \cref{sec:real-data}. Additionally, implementing gradient clipping with Beta divergence loss can enhance computational performance.

\end{Example}

%%%%%%%%%%%%%
\section{Numerical Studies}\label{sec:simulation}
%%%%%%%%%%%%%

In this section, we evaluate the numerical performance of the proposed algorithms. For a randomly drawn element denoted as $x$, we will use $\hat x$ to denote its estimated value, and $\tilde{x}$ to denote a simulated version. \cref{algorithm_RKHS} is referred to as RKHS-TD, \cref{algorithm_RKHS_sketched} is referred to as S-RKHS-TD, \cref{algorithm_PGD} is referred to as GRKHS-TD, and \cref{algorithm_PSGD} is referred to as S-GRKHS-TD. Sampled indices used in the sketching algorithms are denoted as $\hat{N} \subseteq {[n]}, \hat{J} \subseteq [p]$ and $\hat{T_i} \subseteq T_i$ for $i\in \hat{N}$. The codes for all numerical studies are available at \url{https://github.com/RunshiTang/Experiments-for-Tensor-Decomposition-with-Unaligned-Observations}. The RKHS-TD, S-RKHS-TD, GRKHS-TD, and S-GRKHS-TD are implemented in Python 3.11. 
The experiment environment uses 10 CPU cores of AMD EPYC 7763 64-Core Processor with 16 GB memory on Slurm of the Social Science Computing Cooperative at UW-Madison.

%%%%%%%%%%%%%
\subsection{Tensor Decomposition with Unaligned Observations and $\ell_2$ Loss}\label{section_rkhs_simulation}
%%%%%%%%%%%%%

In this section, we apply RKHS-TD and S-RKHS-TD to a simulated dataset with Gaussian noise. 

\paragraph{Generation.}
We choose $D = 251, l = 8, u = 20, R = 5, n = 60$, $\sigma^2=1$, and $p = 51$. For all $r\in \{1, \ldots, R\}$, we first generate $a_r\in \RR^n$, $b_r\in\RR^p$ with elements i.i.d. distributed as uniform$(0,1)$. 
We sample $T$ as a size-$D$ subset of $\{1/739, 2/739, \ldots, 1\}$. 
We generate $\xi_r$ from orthonormal basis functions $\left\{u_i(s)\right\}_{i=1}^{10} \subseteq\mathcal{L}^2([0,1])$. Following the simulation setting of \cite{yuan2010reproducing}, we set $u_1(s)=1$ and $u_i(s)=\sqrt{2} \cos ((i-1) \pi s)$ for $i=2, \ldots, 10$. We generate $x_{r, i} \sim \operatorname{Unif}[-1 / i, 1 / i]$ independently, and let $\xi_r(\cdot)=\sum_{i=1}^{10} x_{r, i} u_i(\cdot)$ (An instance of $\xi_1,\ldots, \xi_R$ is visualized in  \cref{fig_xi_ls} in Appendix). 
Finally, we define the low-rank functional tensor $\widetilde{\mathbf{\Lambda}}_{ij}(t) = \sum_{r=1}^R 10\sqrt{r} \cdot (a_r)_i\cdot (b_r)_j\cdot \xi_r(t)$.

To generate the observed data from $\widetilde{\mathbf{\Lambda}}_{ij}(t)$, we first sample $T$ as a size-$D$ subset of $\{1/739, 2/739, \ldots, 1\}$. Then, we sample $\{d_i\}_{i = 1}^n$ i.i.d. uniformly from $\{l, l+1, \ldots, u-1, u\}$ and sample subsets $\{T_i\}_{i = 1}^n$ with size $d_i$ uniformly from $T$. Thus, we obtained $\widetilde{\mathbf{\Lambda}}_{ij}(t), t\in T_i.$ To simulate the dataset, we generate
\begin{equation}\label{eq_generate_x}
	\widetilde{\bcX}_{ij}(t) \sim \operatorname{Normal}\left(\widetilde{\mathbf{\Lambda}}_{ij}(t), \sigma^2\right); \quad i\in \{1,\ldots, n\}, j\in\{1, \ldots, p\}, t\in T_i.
\end{equation}

\paragraph{Decomposition.}

Next, we apply the tensor decomposition with unaligned observations on $\widetilde{\bcX}_{ij}(t)$ and denote the output as $\hat{\mathbf{\Lambda}}_{ij}(t)$ with the Bernoulli kernel \eqref{eq_Bernoulli_ker}. 
In RKHS-TD (\cref{algorithm_RKHS}) and S-RKHS-TD (\cref{algorithm_RKHS_sketched}), we set penalty coefficient $\lambda^\prime = 10^{-4}$ and target rank $R = 5$. In S-RKHS-TD, we use sketching size $s_2 = 40, s_3 = 10$ and varying $s_1$. 
To measure the performance, we introduce the following measure of goodness of fit for the outcome of the $h$th iteration:
\begin{equation}\label{eq_fit}
    \begin{split}
    & \operatorname{Relative Loss}_h = \frac{\sum_{i=1}^n\sum_{j=1}^p \sum_{t\in T_i}\left(\bcX_{ij}(t) - \hat \bcX_{ij}^h(t)\right)^2}{\sum_{i=1}^n\sum_{j=1}^p \sum_{t\in T_i}\left(\bcX_{ij}(t)\right)^2}, \quad h=1,2,\ldots\\
    & \text{where} \quad \hat \bcX_{ij}^h(t) = \sum_{r=1}^R (\hat a_r)_i\cdot (\hat b_r)_j\cdot \hat \xi_r(t), t\in T_{i} \text{ is the estimation at the $h$th iteration}.
    \end{split}
\end{equation}

\begin{figure}[htbp]
	\centering
        \begin{subfigure}{0.5\textwidth}
		\includegraphics[width=\textwidth]{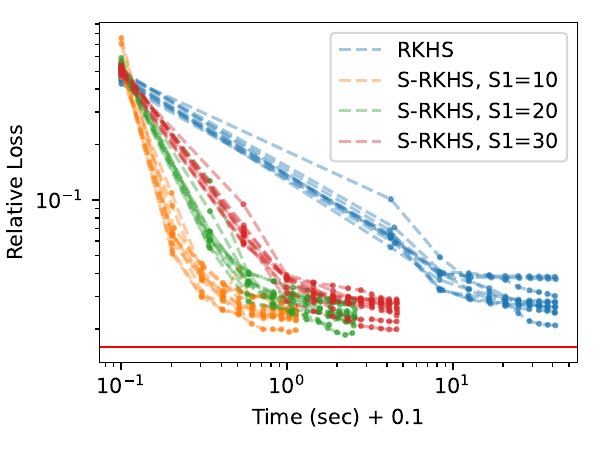}
		\caption{}
	\end{subfigure}
	\begin{subfigure}{0.37\textwidth}
		\includegraphics[width=\textwidth]{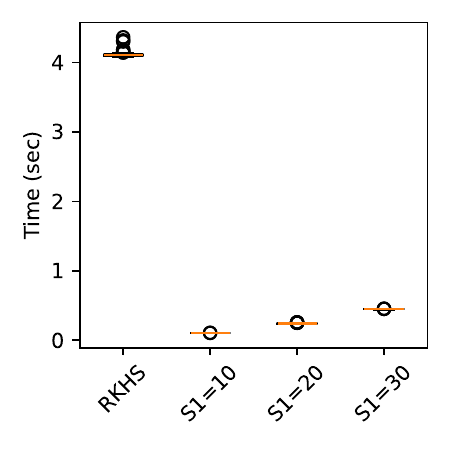}
		\caption{}
	\end{subfigure}
	\hfill
	\begin{subfigure}{.85\textwidth}
		\includegraphics[width=\textwidth]{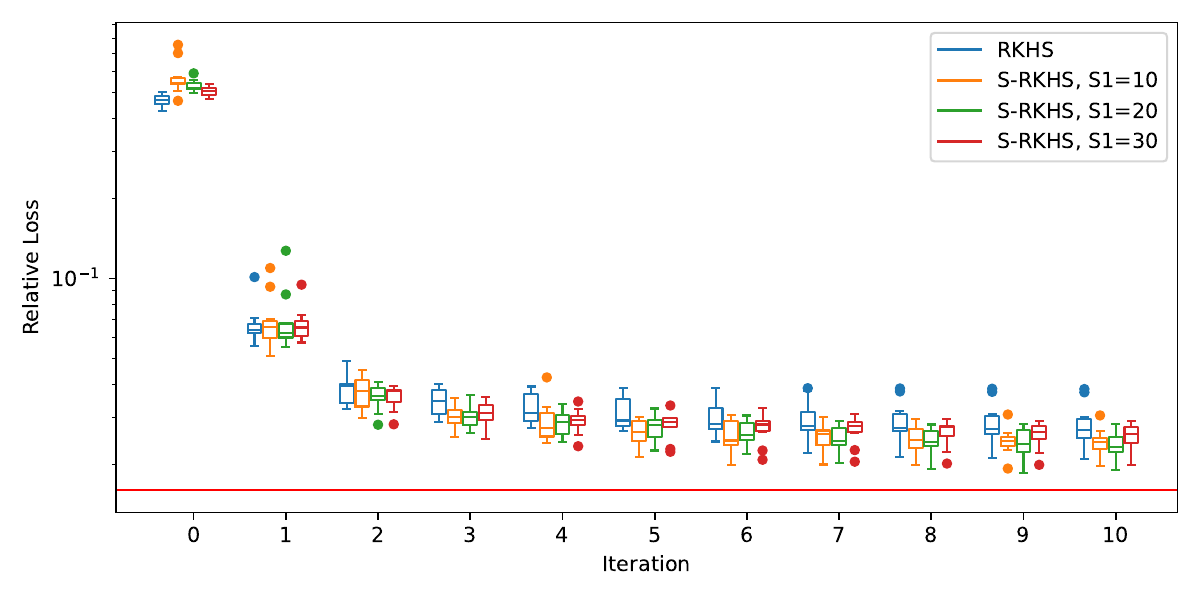}
		\caption{}
	\end{subfigure}
	\caption{Simulation results are presented for RKHS-TD and S-RKHS-TD with a target rank $R = 5$. In (a) and (c), four different algorithms are represented by distinct colors. 
    In (a), each algorithm is depicted with 10 dashed lines, representing 10 simulation trajectories. For each simulation, we record time cost (x-axis, plus 0.1 for log-scale) and relative loss (y-axis, defined in \eqref{eq_fit}) for initialization (Time $= 0$) and the subsequent 10 iterations. The dashed horizontal line indicates the nominal relative loss: $\operatorname{RelativeLoss}(\widetilde{\mathbf{\Lambda}}) \approx 0.017$; (b) displays a box plot illustrating the time cost per iteration for different algorithms. The x-ticks `S1=$X$' correspond to S-RKHS-TD with $s_1 = X$; (c) shows a box plot for the relative loss of each algorithm at initialization and during the first 10 iterations. Iteration $=0$ corresponds to the initialization stage.} 
	\label{fig_RKHS_setting1_R5}
\end{figure}

The results are plotted in \cref{fig_RKHS_setting1_R5}. The algorithms reach the lowest possible relative error they can reach in a few iterations. A similar phenomenon is also observed in High Order Orthogonal Iteration for tensor tucker decomposition and Alternative Least Squares for tabular tensor CP decomposition. From panel (c) of \cref{fig_RKHS_setting1_R5}, we can see that as $s_1$ increases, the median relative error decreases with each iteration in S-RKHS-TD and RKHS-TD has the lowest median relative error. However, each iteration takes longer, as shown in panel (b). Panel (a) shows that RKHS-TD takes much longer to achieve the same level of relative error compared to S-RKHS-TD. All S-RKHS-TD nearly complete all 10 iterations when RKHS completes its first iteration. 

{We also examine the influence of the hyperparameter $\lambda^\prime$ in RKHS-TD. For each setting, we perform 10 simulations, each consisting of 10 iterations. The mean and standard deviation of the relative loss at the final iteration are presented in \cref{table_rkhs_lambda}. The results indicate no significant impact of $\lambda^\prime$ on the performance while a smaller $\lambda^\prime$ leads to a slightly better relative loss. This is unsurprising as in \eqref{loss_xi_2_regularized}, a smaller $\lambda^\prime$ implies the loss is closer to the least square loss. 

\begin{table}[!ht]
    \centering
    \begin{tabular}{l|lllll}
    \hline
        $\lambda^\prime$ & 0.00001 & 0.00005 & 0.0001 & 0.0005 & 0.001 \\ \hline
        Mean$(\operatorname{RelativeError})$ & 0.02478 & 0.02531 & 0.02731 & 0.02632 & 0.02844 \\
        SD$(\operatorname{RelativeError})$ & 0.00318 & 0.00385 & 0.00375 & 0.00214 & 0.00424 \\ \hline

    \end{tabular}
    \caption{Mean and standard deviation of the relative loss in the last iterations for different $\lambda^\prime$ used in RKHS-TD}
    \label{table_rkhs_lambda}
\end{table}}

%%%%%%%%%%%%%
\subsection{Tensor Decomposition with Unaligned Observations and Poisson Loss}\label{section_rkhs_simulation_grkhs}
%%%%%%%%%%%%%

In this section, we apply GRKHS-TD and S-GRKHS-TD with Poisson loss to a simulated dataset from Poisson distribution.

\paragraph{Generation.}
We set $D = 251, l = 8, u = 20, n = 60, p = 51, \delta = 10^{-10}$ and $R=5$. 
For all $r\in \{1, \ldots, R\}$, we first generate $a_r\in \RR^n$, $b_r\in\RR^p$ with elements i.i.d. distributed as uniform$(0,1)$. 
We sample $T$ as a size-$D$ subset of $\{1/739, 2/739, \ldots, 1\}$. 
We generate $\xi_r$ from orthonormal basis functions $\left\{u_i(s)\right\}_{i=1}^{10} \subseteq\mathcal{L}^2([0,1])$. Following \cite{yuan2010reproducing}, we set $u_1(s)=1$ and $u_i(s)=\sqrt{2} \cos ((i-1) \pi s)$ for $i=2, \ldots, 10$. We generate $x_{r, i} \sim \operatorname{Unif}[-1 / i, 1 / i]$ independently, and let $\xi_r^*(\cdot)=\sum_{i=1}^{10} x_{r, i} u_i(\cdot)$. We define $\xi_r = \xi_r^* + 1$ to ensure positivity of $\xi_r$'s. 
Finally, we define the low-rank functional tensor $\widetilde{\mathbf{\Lambda}}_{ij}(t) = \sum_{r=1}^R 10\sqrt{r} \cdot (a_r)_i\cdot (b_r)_j\cdot \xi_r(t)$. 

To generate the observed data from $\widetilde{\mathbf{\Lambda}}_{ij}(t)$, we sample $\{d_i\}_{i = 1}^n$ i.i.d. and uniformly from $\{l, l+1, \ldots, u-1, u\}$ and sample subsets $\{T_i\}_{i = 1}^n$ with size $d_i$ uniformly from $T$. Thus, we obtained $\widetilde{\mathbf{\Lambda}}_{ij}(t), \quad t\in T_i.$ To simulate the dataset, we generate 
\begin{equation}\label{eq_generate_y}
	\widetilde{\bcX}_{ij}(t) \sim \operatorname{Poisson}\left(\widetilde{\mathbf{\Lambda}}_{ij}(t) + \delta\right); \quad i\in \{1,\ldots, n\}, j\in\{1, \ldots, p\}, t\in T_i.
\end{equation}

\paragraph{Decomposition.}

Next, we apply the generalized tensor decomposition with unaligned observations to $\widetilde{\bcX}_{ij}(t)$ with Poisson loss. We similarly denote the low-rank tensor yielded by our algorithms as $\hat{\mathbf{\Lambda}}_{ij}(t)$. 
The loss function we used, in this case, is
\be \label{eq_loss_Generalized_RKHS_poisson}
\operatorname{loss}(\hat{\mathbf{\Lambda}}) = \frac{1}{|\Omega|}\sum_{i=1}^n\sum_{j=1}^p \sum_{t\in T_i} \left(\hat{\mathbf{\Lambda}}_{ij}(t) + \delta - \widetilde{\bcX}_{ij}(t) \log \left(\hat{\mathbf{\Lambda}}_{ij}(t) + \delta\right)\right). 
\ee

\begin{figure}[!ht]
	\centering
	\begin{subfigure}{0.5\textwidth}
		\includegraphics[width=\textwidth]{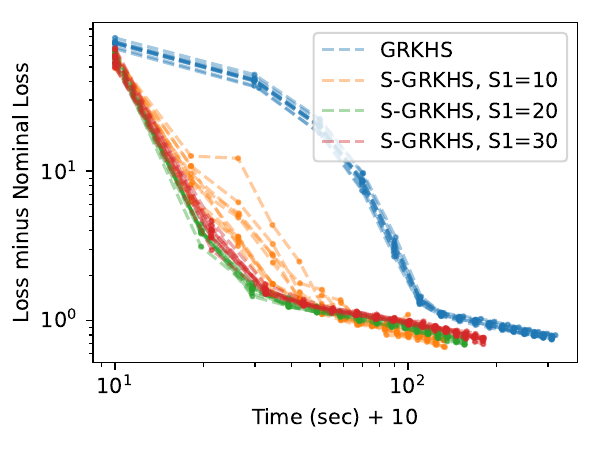}
		\caption{}
	\end{subfigure}
	\begin{subfigure}{0.37\textwidth}
		\includegraphics[width=\textwidth]{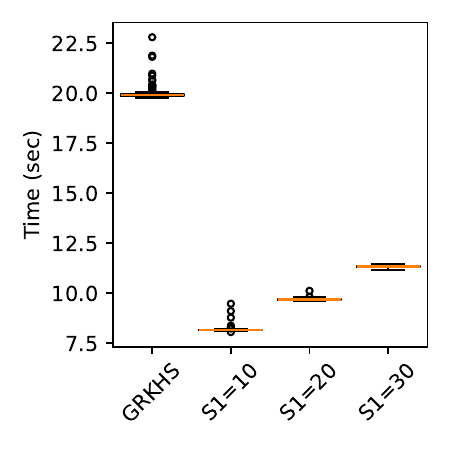}
		\caption{}
	\end{subfigure}
	\hfill
	\begin{subfigure}{0.85\textwidth}
		\includegraphics[width=\textwidth]{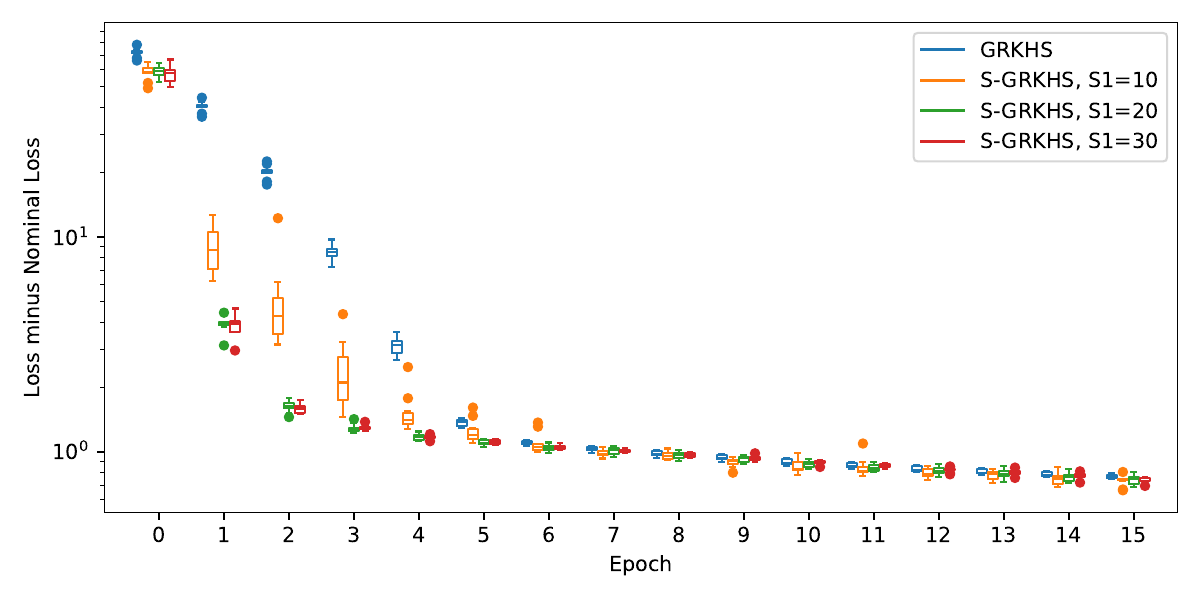}
		\caption{}
	\end{subfigure}
	\caption{Simulation result of GRKHS-TD and S-GRKHS-TD with Poisson loss. The target rank $R = 5$ (same as true rank). In (a) and (c), there are 4 colors representing 4 different algorithms and the y-axis is the loss (defined in \eqref{eq_loss_Generalized_RKHS_poisson}) minus nominal loss $\operatorname{loss}(\widetilde{\mathbf{\Lambda}}) \approx -53.21$. In (a), there are 10 dashed lines for each algorithm representing 10 simulation trajectories. For each simulation, we record the time cost (x-axis, plus 10 for log-scale) and loss for initialization and 1-15 epochs, and each epoch consists of 10 iterations. (b) is the box plot for time cost per epoch for different algorithms. The x-ticks `S1=$X$' corresponds to S-GRKHS-TD with $s_1 = X$. (c) is the boxplot for the loss of each Algorithm at initialization and each epoch. }
	\label{fig_GRKHS_poisson_setting1_R2}
\end{figure}

In GRKHS-TD and S-GRKHS-TD, we use the radial kernel $\mathbb{K}(s,t) = \exp(-|s-t|^2)$ and set the target rank $R=5$, the learning rate $\alpha = 0.4$, and the feasible set $\Phi = \{(a_r,b_r,\theta_r): a_r \geq 0, b_r \geq 0, \theta_{r} \geq 0, \|a_r\| \|b_r\| \|\xi_r\|_\mathcal{H} \leq 10000\}$ for $r = 1, \ldots, R$, where $\theta_r = [\theta_{r, 1}, \ldots, \theta_{r, |T|}]\T$ is the coefficients in $\theta$ involved in the representation of $\xi_r$. We use sketching size $s_2 = 20, s_3 = 10$ and varying $s_1$. We use the gradient clipping discussed in \cref{example_poisson} with $c = 0.5$. 
We evaluate the goodness of fit by calculating the loss function $F$ at $t$th iteration using the full sample and denote it as $\operatorname{loss}_t$.
Note that in large-scale settings, computing the exact value of $\hat \bcX$ and $\operatorname{loss}_t$ in every iteration can be time-consuming. To this end, we introduce epochs, where multiple iterations are executed per epoch, and we calculate $\hat \bcX$ and $\operatorname{loss}_h$ at the end of each epoch, rather than after each iteration.  
The simulation result is shown in \cref{fig_GRKHS_poisson_setting1_R2}. The nominal loss $\operatorname{loss}(\widetilde{\mathbf{\Lambda}}) \approx -53.21$ and all settings are approaching the nominal loss. 

{We also examine the influence of the hyperparameter $\lambda$ in G-RKHS-TD. For each setting, we perform 10 simulations, each consisting of 15 epochs and each epoch consisting of 10 iterations. The mean and standard deviation of the loss at the final iteration are presented in \cref{table_grkhs_lambda}. The results indicate no significant impact of $\lambda$ on the performance. 

\begin{table}[!ht]
    \centering
    \begin{tabular}{l|lllll}
    \hline
        $\lambda$ & 1000 & 5000 & 10000 & 50000 & 100000 \\ \hline
        Mean$(\operatorname{loss})$  & -52.7673297 & -52.43709434 & -52.43664718 & -52.44176774 & -52.44900041 \\ 
        SD$(\operatorname{loss})$ & 0.008327904 & 0.022194256 & 0.01714792 & 0.017745255 & 0.015141073 \\ \hline
    \end{tabular}
    \caption{Mean and standard deviation of the last iteration for different $\lambda$ used in RKHS-TD}
    \label{table_grkhs_lambda}
\end{table}}

%%%%%%%%%%%%%
\section{Real Data Experiments}\label{sec:real-data}
%%%%%%%%%%%%%

In this section, we evaluate the performance of the following algorithms: RKHS-TD, S-RKHS-TD, GRKHS-TD, and S-GRKHS-TD, using the real-world dataset known as Early Childhood Antibiotics and the Microbiome (ECAM) \cite{bokulich2016antibiotics}.
The ECAM dataset includes 42 infants with multiple fecal microbiome measurements from birth over the first 2 years of life. 
Among the 42 infants, 30 were dominantly ($> 50\%$ of feedings) breastfed for the first 3 months and 12 were dominantly formula-fed. 
The fecal microbiome of each infant was sampled on different days across different infants.
Suppose we are interested in the counts of 50 bacterial genera in the fecal microbiome sample of each infant at different times. The observed data can be organized as $\bcY_{ij}(t)\in \mathbb{N}, t\in T_i,$ where $i\in [n]$ denote different infants, $j=[p]$ denote different bacterial genera, $t \in T_i$ denotes the age of days at sampling time, and $T_i$ denotes the set of all sampling time points of infant $i$. 
So the latent (unobserved) counts of bacterial genera $j$ of infant $i$ at $t \notin T_i$ with the observed data can be represented as an order-3 functional tensor $\bcY_{ij}(t)\in \mathbb{N}$. 
The original ECAM data is count-valued, so we apply the centered-log-ratio (CLR) transformation \cite{aitchison1986statistical,shi2022high} to transform the data as: $\bcX_{ij}(t) = \log\left\{(\bcY_{ij}(t)+0.5) / (\sum_{j' = 1}^p(\bcY_{ij'}(t)+0.5))\right\}$. 

\begin{figure}[!ht]
    \centering
    \begin{subfigure}{0.35\textwidth}
        \includegraphics[width=\textwidth]{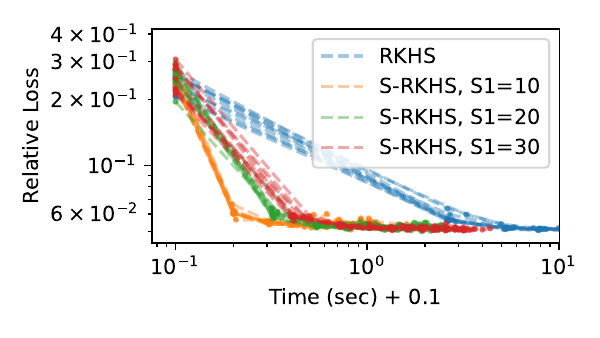}
        \caption{$R=4, s_2=20$}
    \end{subfigure}
        \begin{subfigure}{0.35\textwidth}
        \includegraphics[width=\textwidth]{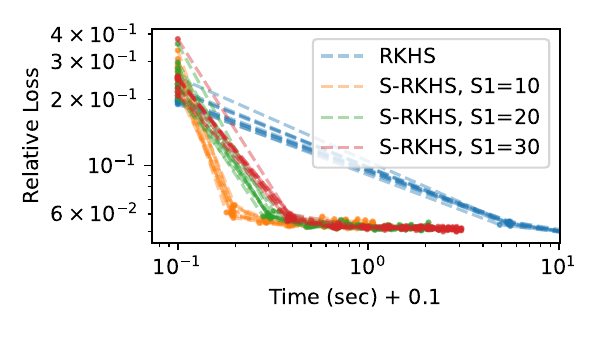}
        \caption{$R=6, s_2=20$}
    \end{subfigure}\\
    \begin{subfigure}{0.35\textwidth}
        \includegraphics[width=\textwidth]{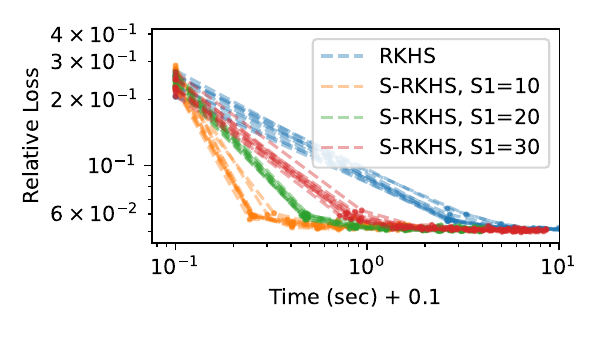}
        \caption{$R=4, s_2=40$}
    \end{subfigure}
        \begin{subfigure}{0.35\textwidth}
        \includegraphics[width=\textwidth]{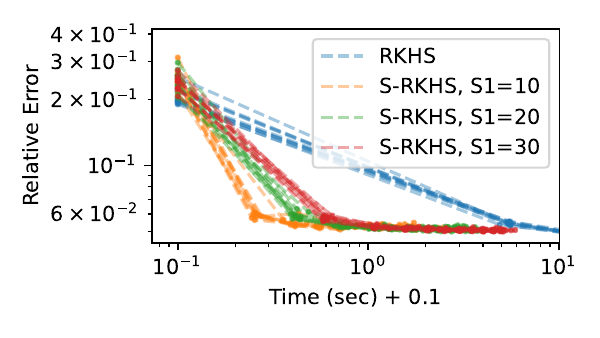}
        \caption{$R=6, s_2=40$}
    \end{subfigure}
    \caption{Comparison of RKHS-TD and S-RKHS-TD in real data experiment on ECAM dataset. There are 10 dashed lines for each Algorithm representing 10 simulation trajectories. For each simulation, we record the time cost (x-axis, plus 0.1 for log-scale) and relative loss (y-axis, defined in \eqref{eq_fit}) for initialization and 1-15 iterations and display the first 10 seconds. } 
    \label{fig_RKHS_realdata_comparison}
\end{figure}

Then, we apply RKHS-TD and S-RKHS-TD with the Bernoulli kernel \eqref{eq_Bernoulli_ker}, penalty coefficient $\lambda^\prime = 10^{-4}$ and $s_3 = 10$, while varying $s_1, s_2$, and $R$. The results are visualized in \cref{fig_RKHS_realdata_comparison}. It is noteworthy that nearly all trajectories converge effectively. Particularly, S-RKHS-TD exhibits significantly faster convergence compared to RKHS-TD, especially when $s_1$ is small. 

\begin{figure}[ht!]
	\centering
	\begin{subfigure}{0.35\textwidth}
		\includegraphics[width=\textwidth]{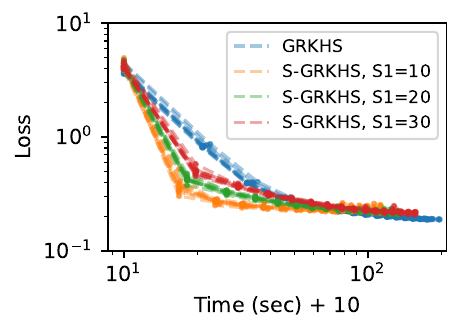}
		\caption{$R=4, s_2=20$}
	\end{subfigure}
	\begin{subfigure}{0.35\textwidth}
		\includegraphics[width=\textwidth]{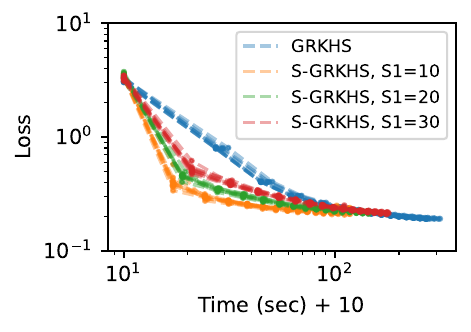}
		\caption{$R=6, s_2=20$}
	\end{subfigure}
	\begin{subfigure}{0.35\textwidth}
		\includegraphics[width=\textwidth]{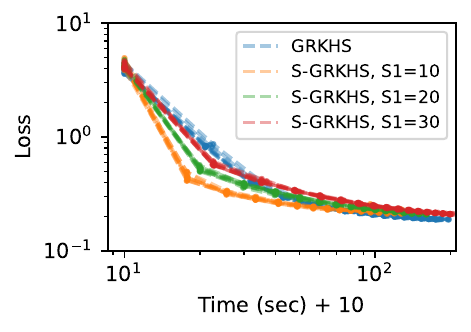}
		\caption{$R=4, s_2=40$}
	\end{subfigure}
	\begin{subfigure}{0.35\textwidth}
		\includegraphics[width=\textwidth]{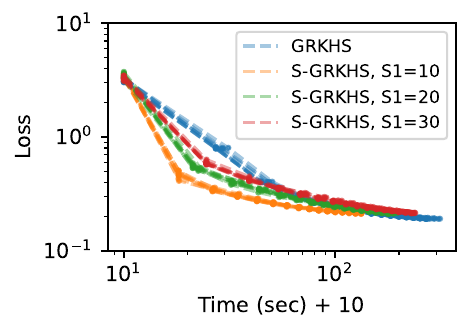}
		\caption{$R=6, s_2=40$}
	\end{subfigure}
\caption{Comparative analysis of GRKHS-TD and S-GRKHS-TD using the Beta divergence loss (defined in \eqref{eq:beta-divergence}) in real data experiment on ECAM dataset. There are 10 dashed lines for each Algorithm representing 10 simulation trajectories. For each simulation, we record the time cost (x-axis, plus 10 for log-scale) and loss (y-axis) for initialization and 1-15 epochs and each epoch consists of 10 iterations. }
\label{fig_GRKHS_realdata_beta_d_comparison}
\end{figure}

Next, we transform the ECAM data to the relative abundance $\bcX_{ij}(t) = \bcY_{ij}(t) / (\sum_{j' = 1}^p\bcY_{ij'}(t))$ and apply GRKHS-TD and S-GRKHS-TD with the radial kernel $\mathbb{K}(s,t) = \exp(-|s-t|^2)$ and Beta divergence loss (see discussions in \cref{example_beta_d}) with $\beta = 0.5$. 
We define $\Phi = \{(a_r,b_r,\theta_r): a_r \geq 0, b_r \geq 0, \theta_{r} \geq 0, \|a_r\| \|b_r\| \|\xi_r\|_\mathcal{H} \leq 10000\}$ for $r = 1, \ldots, R$. We set the learning rate $\alpha$ to 0.1 and vary the target rank $R$. We use a sketching size of $s_3 = 8$ and varying $s_1$ and $s_2$. 
To address the gradient-related issues discussed in \cref{example_beta_d}, we use the $f_\delta$ introduced in \cref{example_beta_d} with $\delta = 10^{-6}$ and implement the gradient clipping with $c = 1$. Our training process follows an epoch-based approach as outlined in \cref{section_rkhs_simulation_grkhs}, where each epoch comprises 10 iterations. The results of our experiments are plotted in \cref{fig_GRKHS_realdata_beta_d_comparison}. Note that though the data is sparse, the loss of trivial estimation $\hat \bcY_{ij}(t) \equiv 0$ is approximately 4000, which is much higher than the ones achieved by our algorithms. 

Finally, we demonstrate the practicality of the proposed S-RKHS-TD and S-GRKHS-TD (with Beta divergence loss and $s_2 = 20$) between breast-fed (bd) and formula-fed (fd) infants. 

{
To mimic the use in practice, we introduce the following stop criteria. 
We terminate the S-RKHS-TD iteration when:
\begin{equation}\label{eq_stop_criterion}
    1 = \prod_{g = 1}^{h_0} I({\{\operatorname{RelativeLoss}_{h-g+1} > \operatorname{RelativeLoss}_{h-g} + \varepsilon\}}), 
\end{equation}
where $I(\cdot)$ denotes the indicator function, $\varepsilon \geq 0$ is a predefined threshold, and $h_0$ is a predefined integer. If \eqref{eq_stop_criterion} is satisfied, it indicates that the relative loss improvement over the last $h_0$ iterations has been consistently less than $\varepsilon$. 
Similarly, we propose the following similar stopping criterion for S-GRKHS-TD: 
\begin{equation}\label{eq_stop_criterion_grkhs}
	1 = \prod_{g = 1}^{h_0} I({\{\operatorname{loss}_{t-g+1} > \operatorname{loss}_{t-g} - \varepsilon\}}). 
\end{equation}
If the criterion is met at the $h$th iteration, we set the result from the $h-h_0$ iteration as the final output. }

\begin{figure}[!ht]
    \centering
        \includegraphics[width=0.9\textwidth]{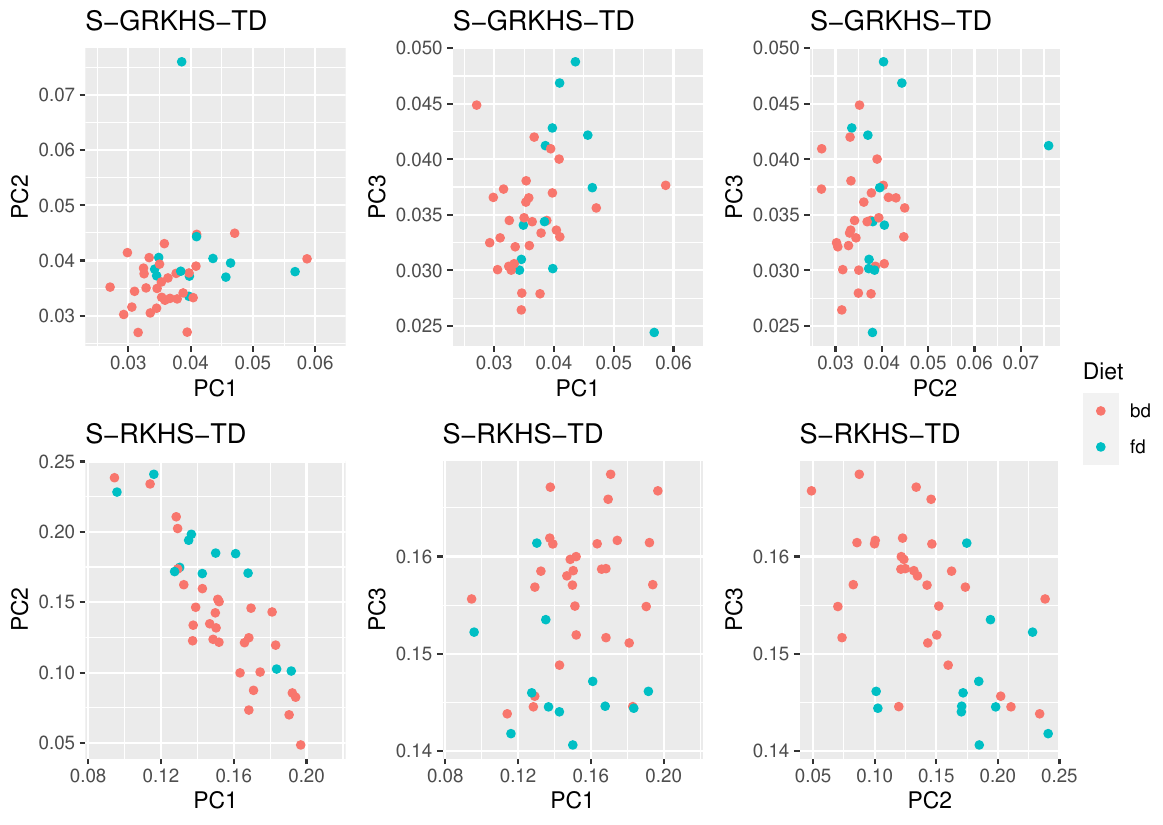}
        \caption{Loading $A$ in S-RKHS-TD and S-GRKHS-TD, $R = 3$}
    \label{fig_ECAM_loadings}
\end{figure}

For comparison, we also discretize the time mode of the ECAM dataset after CLR or relative abundance transformation to obtain tabular tensors; then we apply the functional tensor singular value decomposition (tabular FTSVD\footnote{The codes implementing FTSVD are sourced from \url{https://github.com/Rungang/functional_tensor_svd}.}) \cite{han2021guaranteed} and standard CP decomposition\footnote{The codes implementing CP decomposition are sourced from \url{https://cran.r-project.org/web/packages/rTensor/}.} \cite{kolda2009tensor} to the tabular tensor with CLR transformation, and apply the standard GCP decomposition\footnote{The codes implementing GCP decomposition are sourced from \url{https://www.tensortoolbox.org/gcp_opt_doc.html}.} with Beta divergence loss \cite{hong2020generalized} to the tabular tensor with relative abundance transformation. The hyperparameter $\beta$ in standard GCP and S-GRKHS-TD is chosen to be 0.5. 
We computed the Silhouette score\footnote{A higher Silhouette score indicates more effective data clustering.} \cite{rousseeuw1987silhouettes}to evaluate the clustering performance of CP, FTSVD, GCP, RKHS-TD, and GRKHS-TD on bd and fd infants. 
This was done by computing $k$-means clustering with $k = 2$ on the output loading matrix (i.e., the output $A$ of S-RKHS and S-GRKHS) and comparing the clusters to the infants' true label (bd or fd). Notably, S-RKHS-TD and S-GRKHS-TD recorded the highest Silhouette scores with rank $R = 3$, outperforming CP (with rank $R=6$) and FTSVD (with rank $R=4$), likely due to their ability to avoid discretization and thus retain more information that could be lost during tabular tensor data preprocessing. Additionally, we present the loading estimates of S-RKHS-TD and S-GRKHS-TD, both at rank $R=3$, in \cref{fig_ECAM_loadings}. 

\begin{figure}[!ht]
    \centering
        \includegraphics[width=0.7\textwidth]{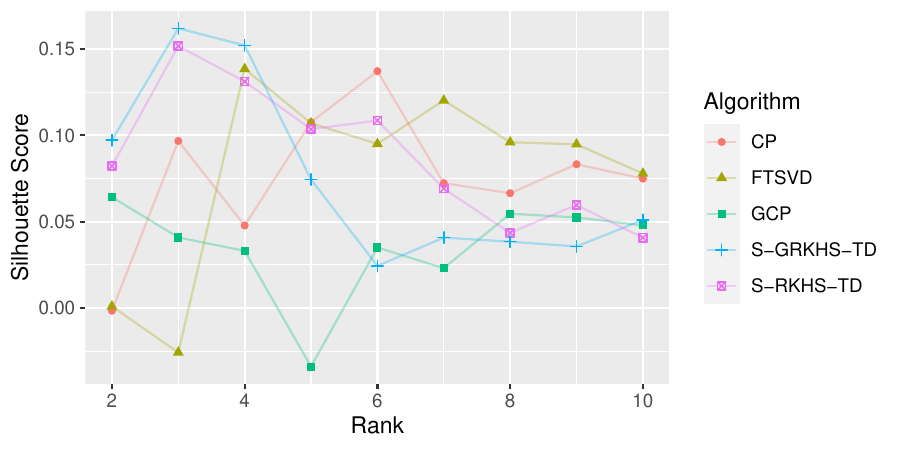}
    \caption{Silhouette scores by different algorithms with different ranks.}
    \label{fig_ECAM_comparison_s}
\end{figure}

%%%%%%%%%%%%%
\section{Discussions}\label{sec:discussions}
%%%%%%%%%%%%%

This paper introduced a framework for tensor decomposition that handles tensors with a mode containing unaligned observations (a functional mode) by decomposing them into a sum of rank-one tensors. The unaligned mode is represented using functions in a reproducing kernel Hilbert space (RKHS), providing a flexible and robust data representation. A versatile loss function was developed, capable of effectively handling various types of data, including binary, integer-valued, and positive-valued types. To compute these tensor decompositions with unaligned observations, we proposed the algorithms RKHS-TD and GRKHS-TD. Additionally, we implemented a stochastic gradient method, S-GRKHS-TD, to enhance computational efficiency. For scenarios where the $\ell_2$ loss function is employed, we introduced a sketching algorithm, S-RKHS-TD, to further improve efficiency.

We emphasize the important role of the RKHS framework and the Representer Theorem in our formulation. In~\eqref{optimization_problem}, each $\xi_r$ is assumed to lie in some RKHS (RKHS), which is a highly general class of functions that may be spanned by infinitely many basis functions. The Representer Theorem implies that the solution to the optimization problem admits a finite-dimensional representation, as shown in~\eqref{loss_xi_representation}, once the time points are observed. This allows the optimization to be carried out over finitely many parameters, making computation feasible despite the infinite-dimensional nature of the function space. 
An alternative approach would be to directly assume that each $\xi_r$ satisfies the form in~\eqref{loss_xi_representation}, for example by choosing a finite-degree polynomial basis or truncated Fourier basis. However, such a restriction would limit the expressiveness and flexibility of the model.

In the simulation study for S-GRKHS-TD, we observed that a smaller learning rate can lead to better ultimate minimization results. However, it may also result in slower progress during the initial epochs, leading to a slower convergence rate. To achieve a balance between the need for a larger learning rate at the beginning and a smaller one towards the end, a gradient decay strategy can be employed in S-GRKHS-TD. In addition, variations of stochastic gradient descent (e.g., Adam \cite{kingma2014adam}) may also be applied.

While the primary emphasis of this paper lies in the decomposition of tensors with two tabular modes and one functional mode, the proposed methods can be extended to handle more general tensors with varying numbers of tabular modes and/or functional modes. We pursue this further in the forthcoming paper \cite{HIFI-2024}, where we consider any combination of tabular (finite-dimensional) and functional (infinite-dimensional) modes. It is also interesting to extend our framework to cover other tensor decomposition methods, such as Tucker decomposition \cite{hitchcock1927expression}, tensor-train decomposition \cite{zhou2022optimal}, and tensor network model \cite{ye2018tensor}. These extensions are to be investigated in future work.

\subsection*{Acknowledgements}
A.R.Z. thanks Zhaoran Wang for the opportunity to present at the Departmental Colloquium of the Department of Industrial Engineering \& Management Sciences at Northwestern University in October 2022. This event facilitated a valuable meeting between T.K. and A.R.Z. Additionally, A.R.Z. appreciates the insightful feedback received from the audience during this colloquium. The research A.R.Z. was partially funded by the NSF Grant CAREER-2203741.

\bibliographystyle{plain}
\bibliography{reference}
\clearpage

\appendix

\begin{center}
\textbf{\Large{}{}{}{}{}{} Supplementary materials}{\Large{}{}}\\
 {\huge{}{} }{\huge\par}
\par\end{center}

\section{More Details of Fast Computation via Sketchings}\label{sec:supplement_sketching}

In this section, we discuss the correspondence between the sketching matrix and sampling procedure discussed in \cref{sec:sketching} and provide concrete algorithm of the sketched version of \cref{algorithm_RKHS}.

First, there is a natural bijective mapping $f: k \mapsto (i,j,t)$ defined by the correspondence between $\bcX$ and its vectorization ${x}$: the $k$th element of ${x}$ corresponds to some entry of $\bcX$, and we denote the index of this entry as $(i,j,t)$. 
Then, for given integers $\hat{n} \leq n, \hat{p} \leq p$, we firstly sample $i_1, \ldots, i_{\hat{n}}$ i.i.d. uniformly from $\{1, \ldots,n\}$ and sample $j_1, \ldots, j_{\hat{p}}$ i.i.d. uniformly from $\{1, \ldots,p\}$. 
Next, for $k = 1, \ldots, \hat{n}$ and given integers $\hat{u}_{i_k}$, we sample $t^{i_k}_1, \ldots, t^{i_k}_{\hat{u}_{i_k}}$ i.i.d. uniformly from $T_{i_k}$. 
Denote $\hat{N} = \{i_h\}_{h = 1}^{\hat{n}}, \hat{J} = \{j_h\}_{h = 1}^{\hat p}$ and $\hat{T}_i = \{ t^{i}_h\}_{h = 1}^{\hat{u}_{i}}$ for $i \in \hat{N}$. Here, $\hat{N}, \hat{J}$, and $\hat{T_i}$ count duplicate elements. 
Finally, we evaluate $\{q_l\}_{l = 1}^{k_\xi} = \{f^{-1}(i,j,t): i \in \hat{N}, j \in \hat{J}, t\in \hat{T_i}\}$ and set $S_{l,q_l} = 1$ for $l = 1,\ldots, k_\xi$ and the other entries of $S$ to be zero. 
Here, $k_\xi = \hat p \sum_{k = 1}^{\hat{n}} \hat{u}_{i_k}$. 

This configuration facilitates the computation of $SD$. Note that $SD = [SD_1, \ldots, SD_R]$. Given the fact that $S$ is a sketching matrix, we further have $SD_r = S[(P_1 a_r) * (P_2 b_r) * (P_3 \mathbb{K}(T, T))] = (SP_1 a_r) * (SP_2 b_r) * (SP_3 \mathbb{K}(T, T))$. The matrices $SP_i$ can be viewed as the permutation matrices for the sampled sub-tensor $\{\bcX_{ij}(t): i \in \hat{N}, j \in \hat J, T_i \in \hat T_i\}$. 
Hence, using the above sampling procedure to generate $S$ allows us to calculate $SD$ by $SD =  \hat D$, where $\hat D$ is obtained by replacing $\{\bcX_{ij}(t): i \in {N}, j \in {J}, t \in {T_i}\}$ with $\{\bcX_{ij}(t): i \in \hat{N}, j \in \hat{J}, t\in \hat{T_i}\}$ in the definition of $D$. A similar argument holds for$S{x}$. 

We summarize the procedure of constructing the sketched matrix $SD$ to \cref{algorithm_sketched_coefficient} and the overall procedure of sketched tensor decomposition with unaligned observations to \cref{algorithm_RKHS_sketched}. 

\begin{algorithm}
	\caption{Sketched Matrix Construction}
 \label{algorithm_sketched_coefficient}
	\begin{algorithmic}[1]
		\Require {Matrices $A,B$ and sets $T_i \subseteq[0, 1]$; 
			randomly sampled subsets  $\hat{N} \subseteq{[n]}, \hat{J} \subseteq [p]$ and $\hat{T_i} \subseteq T_i$ for $i\in \hat{N} $}
		\Ensure $\hat D$
		\State{Let $\hat{T} = \bigcup_{i\in \hat{N}} \hat{T_i}$ and $|\hat \Omega| = \hat{J} \sum_{i\in \hat{N}} |\hat{T_i}|$;}
		\State{Define permutation matrices $\hat P_1 \in \RR^{|\hat \Omega| \times n}, \hat P_2 \in \RR^{|\hat \Omega| \times p}$ and $\hat P_3 \in \RR^{|\hat \Omega| \times |T|}$ such that $(\hat P_1 a_r)_k = (a_r)_{i_k}, (\hat P_2 b_r)_k = (b_r)_{j_k}$ and $(\hat P_3 \xi_r(T))_k = \xi_r(t_k)$ where $i_k, j_k$ and $t_k$ are the indices of the $k$th element of the sampled observations;}
		\State{Let $\hat D_r = [(\hat P_1 a_r) * (\hat P_2 b_r) * (\hat P_3 \mathbb{K}(T, T))]$ for $r\in [R]$;}
		\State{Let $\hat D = [\hat D_1, \ldots, \hat D_R]$;}\\
		\Return{$\hat D$ \label{a_2_prod}}
	\end{algorithmic}
\end{algorithm}

\begin{algorithm}
\caption{Sketched Tensor Decomposition with Unaligned Observations via RKHS (S-RKHS-TD)}
\label{algorithm_RKHS_sketched}
	\begin{algorithmic}[1]
		\Require {Observed functional tensor $\bcX_{ij}(t)$ for $i\in [n]; j=[p]$ and $T_i \subseteq[0, 1]$; 
			Penalty coefficient $\lambda^\prime$;
			Target rank $R$; 
			Maximum iterations $m_{\max}$; Initialization $A, B, \theta$;}
		\Ensure $A, B, \theta$ and $\hat \bcX_{ij}(t)$
		\For{$t$ in $1, \ldots, m_{\max}$}
        \State{Sample subsets $\hat{N} \subseteq{[n]}, \hat{J} \subseteq [p]$ and $\hat{T_i} \subseteq T_i$ for $i\in \hat{N}$ with size $|\hat{N}| = s_1$, $|\hat{J}| = s_2$ and $|\hat{T_i}| = s_3$;}
        \State {Update $A$ by \eqref{loss_a};}
        \State {Update $B$ by \eqref{loss_b};}
        \State {Update $\theta$ by \eqref{eq:loss-function-xi-sketched} where $Sx$ is the vectorized sampled observations and $SD = \hat D$ is calculated by \cref{algorithm_sketched_coefficient};}
		\EndFor\\
		\Return $A, B, \theta$ and $\hat \bcX_{ij}(t)$
	\end{algorithmic}
\end{algorithm}

%%%%%%%%%%%%%%
\paragraph{Time Complexity.}
%%%%%%%%%%%%%%

In \cref{algorithm_sketched_coefficient}, Calculation of $D$ takes $|\hat J| R \sum_{i\in \hat{N}}|\hat{T}_i|$ flops. 
In each iteration of \cref{algorithm_RKHS_sketched}, the update of $A$ and $B$ takes the same flops as in \cref{algorithm_RKHS}. For the functional mode, the cost of sampling is $O(n|\hat N| + p|\hat J| + \sum_{i\in \hat{N}}|\hat{T}_i||T_i|)$, and it takes $O(R^2 |\hat J| |T|^2 \sum_{i\in \hat{N}}|\hat{T}_i|)$ flops to calculate the coefficients in \eqref{eq:loss-function-xi-sketched}. Finally, solving $\theta$ takes $O(R^3|T|^3)$ flops. Thus, assuming $n|\hat N| + p|\hat J| + \sum_{i\in \hat{N}}|\hat{T}_i||T_i| \leq R^2 |\hat J| |T|^2 \sum_{i\in \hat{N}}|\hat{T}_i|$, each iteration in S-RKHS-TD (\cref{algorithm_RKHS_sketched}) takes $O(R^2|T|^2(R|T| + |\hat J| \sum_{i\in \hat{N}}|\hat{T}_i|))$ flops. 

The comparison between \cref{algorithm_RKHS,algorithm_RKHS_sketched}, i.e., the original and sketched tensor decomposition with unaligned observations, is summarized to \cref{table_complexity_RKHS}. Note that both algorithms involve solving $\theta$ by minimizing a quadratic form, which requires $O(R^3|T|^3)$ flops. 

\section{Additional Figures}
\begin{figure}[h]
    \centering
    \includegraphics{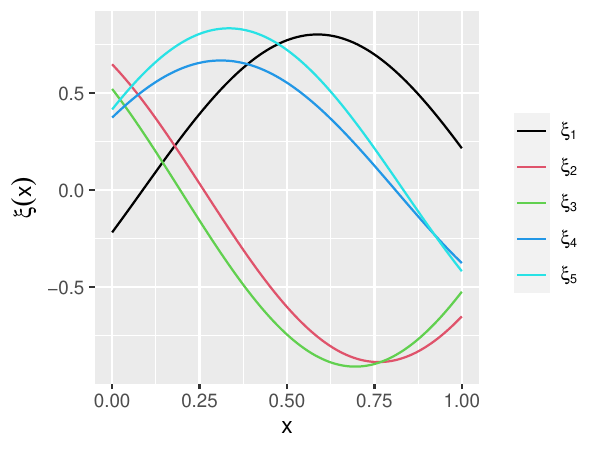}
    \caption{Simulated $\xi$}
    \label{fig_xi_ls}
\end{figure}

%%%%%%%%%%%
\section{Convergence of RKHS-TD}
%%%%%%%%%%%
In this section, we study the RKHS-TD (Algorithm \ref{algorithm_RKHS}) with input $\bcX_{ij}(t), i = 1, \ldots, n, j = 1, \ldots, p, t\in T_i$, $R = 1$, and $\lambda'$. 
We define $a^{(h)},  b^{(h)}$, and $\xi^{(h)}$ as the solution to \eqref{optimization_A}, \eqref{optimization_B}, and \eqref{loss_xi_2} at the $h$th iteration of Algorithm \ref{algorithm_RKHS}. 
Denote the estimation error at the $h$th iteration as
\[
    \varepsilon_h = \max_k\{ |(a^{(h)})_{i_k} - (a)_{i_k}|, |(b^{(h)})_{j_k} - (b)_{j_k}|, |\xi^{(h)}(t_k) - \xi(t_k)|\}.
\]
We denote the vectorized tensor as
\[
    x = \left(\bcX_{11}(T_1)\T, \ldots, \bcX_{1p}(T_1)\T, \ldots, \bcX_{i1}(T_i)\T, \ldots, \bcX_{ip}(T_i)\T , \ldots, \bcX_{n1}(T_n)\T, \ldots, \bcX_{np}(T_n)\T\right)^\top \in \RR^{|\Omega|},
\] 
and
$x_k = \bcX_{i_k, j_k}(t_k), k \in 1,\ldots, |\Omega|$ is the $k$th entry. $K_1(i) = \{k: i_k = i\}, K_2(j) = \{k: k_k = j\}$ and $K_3(t) = \{k: t_k = t\}$ are the maps from one tensor index to all the terms of $x = (x_1, \ldots, x_{|\Omega|})$ that correspond to that tensor index.  
The following theorem establishes the convergence properties of the rank-one RKHS-TD algorithm.
\begin{theorem}
    Consider some given functional tensor $\bcX$ with observation $\bcX_{ij}(t), i = 1, \ldots, n, j = 1, \ldots, p, t\in T_i$. Let $a,b,$ and $\xi$ be the global minimizer in \eqref{optimization_problem} with $R = 1$. 
    Denote $z_{\max} = \max_{i,j,t \in T_i} |\bcX_{ij}(t) - (a)_i (b)_j \xi(t)|$. 
    Assume $\max\{\varepsilon_0, z_{\max}\} \leq c$ and
    \begin{align}
        0<C_1 < \min&\left\{ \sum_{k\in K_1(i)} \xi(t_k)^2 (b)_{j_k}^2, \quad\text{for $\forall i = 1, \ldots, n$}; \right.\notag\\
        &\quad \sum_{k\in K_2(j)} (a)_{i_k}^2 \xi(t_k)^2, \quad\text{for $\forall j = 1, \ldots, p$};\label{eq_assumption_RKHS_TD}\\
        &\quad \left.\sum_{k\in K_3(t)} (a)_{i_k}^2 (b)_{j_k}^2, \quad\text{for $\forall t \in T$};\right\}\notag
    \end{align}
    for some constants $c$ and $C_1$. 
    Then the estimation error $\varepsilon_h$ at the $h$th iteration in Algorithm \ref{algorithm_RKHS} with $\lambda' = 0$ and $R = 1$ satisfies 
    \[
        \varepsilon_{h} \lesssim \max\{C_2\varepsilon_0 - C_3 {z_{\max}}, 0\}^{2^h} + z_{\max}
    \]
    for some constants $C_2$ and $C_3$. Specifically, when $\bcX$ is exactly rank-one, we have $\varepsilon_h$ converges to 0. 
\end{theorem}

\begin{remark}
    The assumption in \eqref{eq_assumption_RKHS_TD} essentially ensures that the objective function is non-singular with respect to each parameter of interest. If $C_1 = 0$, then some term in the minimum must be zero—for example, suppose without loss of generality that $\sum_{k \in K_1(1)} \xi(t_k)^2 (b)_{j_k}^2 = 0$. This implies that for all tensor entries $\bcX_{1j}(t)$, we have $\xi(t)(b)_{j} = 0$. Consequently, the value of $(a)_1$ has no influence on the overall objective function, making $(a)_1$ unidentifiable.
\end{remark}

\begin{proof}
We denote $\bcZ = \bcX - a \circ b\circ \xi$. First, consider the update of $\{\xi_r\}_{r = 1}^R$.
Recall 
\[
      \xi^{(h+1)} = \argmin_{\zeta, r= 1,\ldots, R} \sum_{k=1}^{|\Omega|} \left(x_k - \sum_{r=1}^R (a^{(h)})_{i_k}\cdot (b^{(h)})_{j_k}\cdot \zeta(t_k)\right)^2.
\]
Hence, we have
\[
    \sum_{k=1}^{|\Omega|} \left(x_k - \sum_{r=1}^R (a^{(h)})_{i_k}\cdot (b^{(h)})_{j_k}\cdot \xi^{(h+1)}(t_k)\right)^2 \leq \sum_{k=1}^{|\Omega|} \left(x_k - \sum_{r=1}^R (a^{(h)})_{i_k}\cdot (b^{(h)})_{j_k}\cdot \xi(t_k)\right)^2,
\]
which indicates
\begin{align*}
    &\sum_{k=1}^{|\Omega|} (a^{(h)})_{i_k}^2 (b^{(h)})_{j_k}^2 ( \xi^{(h+1)}(t_k) - \xi(t_k))^2 \\
    \leq & 2 \sum_{k=1}^{|\Omega|} (a^{(h)})_{i_k}  (b^{(h)})_{j_k} ( \xi^{(h+1)}(t_k) - \xi(t_k)) \left( \xi(t_k)   ((a)_{i_k}  (b)_{j_k} - (a^{(h)})_{i_k}  (b^{(h)})_{j_k}) + z_k\right) \\
    \leq & 2 \left| \sum_{k=1}^{|\Omega|} (a^{(h)})_{i_k}  (b^{(h)})_{j_k} ((a)_{i_k}  (b)_{j_k} - (a^{(h)})_{i_k}  (b^{(h)})_{j_k}) \right| \left| \sum_{k=1}^{|\Omega|} \xi(t_k)  ( \xi^{(h+1)}(t_k) - \xi(t_k)) \right| \\
         & + 2 \left| \sum_{k=1}^{|\Omega|} (a^{(h)})_{i_k}  (b^{(h)})_{j_k} \right| \left| \sum_{k=1}^{|\Omega|} z_k  ( \xi^{(h+1)}(t_k) - \xi(t_k)) \right|\\
    \leq & C |\Omega| \max\{|\xi^{(h+1)}(t_k) - \xi(t_k)|\} \left| (ab^{(h)})\T (ab^{(h)} - ab)  \right| + 2 z_{\max} |\Omega|^2 \max\{|\xi^{(h+1)}(t_k) - \xi(t_k)|\}\\
    \leq & C |\Omega| \max\{|\xi^{(h+1)}(t_k) - \xi(t_k)|\} \|ab^{(h)} - ab\|^2 + 2 z_{\max} |\Omega|^2 \max\{|\xi^{(h+1)}(t_k) - \xi(t_k)|\}\\
    \leq & C |\Omega|^2 \max\{|\xi^{(h+1)}(t_k) - \xi(t_k)|\} (\varepsilon_h^2 + z_{\max})
\end{align*}
where $ab^{(h)} = a^{(h)} * b^{(h)}$ and $ab = a * b$. 

Thus, it follows that
\begin{align*}
     C \max \{|\xi^{(h+1)}(t) - \xi(t)|^2\}
    \leq & (C - 3|\Omega|\varepsilon_h^2) \max \{|\xi^{(h+1)}(t) - \xi(t)|^2\} \\
    \leq & \sum_{t \in T}( \xi^{(h+1)}(t) - \xi(t))^2 \sum_{k \in K_3(t)} \left| (a)_{i_k}^2 (b)_{j_k}^2 - 3 \varepsilon_h^2 \right|\\
    \leq & \sum_{t \in T}( \xi^{(h+1)}(t) - \xi(t))^2 \sum_{k \in K_3(t)} (a^{(h)})_{i_k}^2 (b^{(h)})_{j_k}^2  \\
    = &\sum_{k=1}^{|\Omega|} (a^{(h)})_{i_k}^2 (b^{(h)})_{j_k}^2 ( \xi^{(h+1)}(t_k) - \xi(t_k))^2 \\
    \leq & C |\Omega|^2 \max\{|\xi^{(h+1)}(t_k) - \xi(t_k)|\} (\varepsilon_h^2 + z_{\max}), \\
\end{align*}
which yields
\[
    \max_t \{|\xi^{(h+1)}(t) - \xi(t)|\} \lesssim \varepsilon_h^2 + z_{\max}. 
\]

The similar arguments hold for tabular modes. Hence, we have
\[
    \varepsilon_{h+1} \leq C (\varepsilon_h^2 + z_{\max}),
\]
which by  $\varepsilon_h < C {z_{\max}} + c_0$ indicates
\[
    \varepsilon_{h+1} - 2C{z_{\max}} \lesssim (\varepsilon_h - 2C{z_{\max}})^2. 
\]
Thus,  $\varepsilon_{h+1} - 2C{z_{\max}}$ converges to 0 quadratically by the assumption that $\max\{\varepsilon_0, z_{\max}\} \leq c$, and the statements in the theorem follows. 
\end{proof}

\section{Proof of Proposition \ref{proposition_representer_thm}}\label{proposition_representer_thm_proof}
\begin{proof}
    For any fixed $a_r, b_r, r = 1, \ldots, R$ with norm 1 and $\lambda$, 
    by Karush–Kuhn–Tucker optimality condition in Hilbert Space (e.g., Theorem 5.1 in chapter 3 of \cite{ekeland1999convex}) and the convexity of the loss function, there exists some constant $\lambda^\prime$ such that
    the constrained optimization problem 
    \be\label{eq_tmp1}
        \argmin_{\substack{\|\xi_r\|_\mathcal{H} \leq \lambda \\ {r = 1,\ldots,R}}} \frac{1}{|\Omega|}\sum_{i=1}^n\sum_{j=1}^p \sum_{t\in T_i} f\left(\sum_{r=1}^R (a_r)_i\cdot (b_r)_j\cdot \xi_r(t), \bcX_{ij}(t)\right)
    \ee can be replaced by the following regularized problem:
    \[
        \argmin_{\substack{\xi_r \in \mathcal{H} \\ {r = 1,\ldots,R}}} \frac{1}{|\Omega|}\sum_{i=1}^n\sum_{j=1}^p \sum_{t\in T_i} f\left(\sum_{r=1}^R (a_r)_i\cdot (b_r)_j\cdot \xi_r(t), \bcX_{ij}(t)\right) + \lambda^\prime \|\xi_r\|_\mathcal{H}. 
    \]
   In \eqref{eq_tmp1}, the feasible set is bounded and closed, which is also weakly compact by Banach-Alaoglu Theorem \cite{rudin1991functional}. Further note that the loss function is Gateaux-differentiable and convex on $\xi_r$. We claim that a convex Gateaux differentiable map on $\mathcal{H}$ attains its minimum when restricted to any weakly compact subset. 
    Furthermore, any minimizer to the regularized problem admits a form of $\hat{\xi}_r=\sum_{s \in \cup_{i=1}^n T_i} \theta_{r, s} \mathbb{K}(\cdot, s)$ for $\theta=(\theta_{r, s})\in \RR^{|T|}$ by Representer Theorem. 

    We prove the claim. 
    Let $K \subseteq \mathcal{H}$ be a weakly compact subset. Consider a minimizing sequence $\{x_n\} \subset K$ such that $F(x_n) \to \inf_{x \in K} F(x)$. 
    Since $K$ is weakly compact, there exists a subsequence and an element $x^* \in K$ such that $x_{n_i} \to x^*$ weakly in $E$. By convexity, we have:
    \[
    F(x_n) \geq F(x^*) + \frac{\partial F}{\partial (x^* - x_n)}(x^*).
    \]
    Let $n \to \infty$. By the weak continuity of the directional derivative, it follows $\liminf_{n \to \infty} F(x_n) \geq F(x^*)$. 
    Thus it follows that $\inf_{x \in K} F(x) \geq F(x^*)$, and hence $
    F(x^*) = \inf_{x \in K} F(x)$. 
\end{proof}

\end{document}